\documentclass[lettersize,journal]{IEEEtran}

\usepackage{hyperref}
\hypersetup{
  colorlinks   = true,
  urlcolor     = blue,
  linkcolor    = black,
  citecolor   = black
}
\usepackage{graphicx}
\usepackage{subcaption}
\usepackage{amsmath,amsfonts}
\usepackage{algorithmic}
\usepackage{booktabs}
\usepackage{color, colortbl}
\usepackage{dsfont}
\usepackage{microtype}
\usepackage{threeparttable}
\usepackage{soul}
\usepackage{multirow}
\usepackage[linesnumbered,ruled,vlined]{algorithm2e}
\usepackage{chngcntr}

\usepackage[none]{hyphenat}
\usepackage{array}
\usepackage{textcomp}
\usepackage{stfloats}
\usepackage{soul}
\usepackage{verbatim}
\usepackage{cite}
\usepackage{makecell}
\usepackage{float}
\usepackage{multicol}
\usepackage[switch]{lineno}

\usepackage{xspace}
\def\eg{\textit{e.g.}}
\def\ie{\textit{i.e.}}

\def\etal{\textit{et al. }}

\makeatletter
\newcommand*{\rome}[1]{\expandafter\@slowromancap\romannumeral #1@}
\makeatother

\definecolor{orange}{rgb}{1.0, 0.5, 0.0}

\usepackage[normalem]{ulem}
\definecolor{Gray}{gray}{0.9}
\definecolor{Red}{RGB}{245,220,220}

\begin{document}
\title{Unveiling the Tapestry: the Interplay of Generalization and Forgetting in Continual Learning}

\author{{Zenglin Shi*, Jie Jing*, Ying Sun, Joo Hwee Lim, Mengmi Zhang$\dag$

*Equal contribution \quad $\dag$Corresponding author
}
\thanks{ZS is with Nanyang Technological University (NTU) and Agency for Science, Technology and Research (A*STAR), Singapore
 and Hefei University of Technology, China.
JJ is with NTU, A*STAR, and Sichuan University, China.
YS and JL are with A*STAR.
 MZ is with NTU, A*STAR}
\thanks{Address correspondence to mengmi.zhang@ntu.edu.sg}
\thanks{Manuscript received mm-dd-yyyy; revised mm-dd-yyyy.}}

\markboth{Journal of \LaTeX\ Class Files,~Vol.~14, No.~8, August~2021}%
{Shell \MakeLowercase{\textit{et al.}}: A Sample Article Using IEEEtran.cls for IEEE Journals}


\maketitle

\begin{abstract}
In AI, generalization refers to a model's ability to perform well on out-of-distribution data related to the given task, beyond the data it was trained on. For an AI agent to excel, it must also possess the continual learning capability, whereby an agent incrementally learns to perform a sequence of tasks without forgetting the previously acquired knowledge to solve the old tasks.
Intuitively, generalization within a task allows the model to learn underlying features that can readily be applied to novel tasks, facilitating quicker learning and enhanced performance in subsequent tasks within a continual learning framework. Conversely, continual learning methods often include mechanisms to mitigate catastrophic forgetting, ensuring that knowledge from earlier tasks is retained. This preservation of knowledge over tasks plays a role in enhancing generalization for the ongoing task at hand.
Despite the intuitive appeal of the interplay of both abilities, existing literature on continual learning and generalization has proceeded separately. In the preliminary effort to promote studies that bridge both fields, we first present empirical evidence showing that each of these fields has a mutually positive effect on the other. 
Next, building upon this finding, we introduce a simple and effective technique known as Shape-Texture Consistency Regularization (STCR), which caters to continual learning. STCR learns both shape and texture representations for each task, consequently enhancing generalization and thereby mitigating forgetting. Remarkably, extensive experiments validate that our STCR, can be seamlessly integrated with existing continual learning methods, including replay-free approaches. Its performance surpasses these continual learning methods in isolation or when combined with established generalization techniques by a large margin. Our data and source code are available \href{https://github.com/ZhangLab-DeepNeuroCogLab/distillation-style-cnn}{here}.
\end{abstract}


\section{Introduction}
\label{sec:intro}

\IEEEPARstart{A}{rtificial}  intelligent agents must possess the capability not only to learn and recognize out-of-distribution data related to a given task but also to excel at continual learning, acquiring knowledge over a sequence of tasks without forgetting information from earlier tasks. Both these capabilities are crucial for the deployment of AI in dynamically changing and complex environments. Consider an autonomous vehicle journeying from California to Boston in the US, where it encounters varying weather conditions, transitioning from sunny to snowy. Additionally, the vehicle may come across novel objects along the route, requiring continuous learning to detect these new objects without forgetting the previously learned ones. While both generalization and continual learning are essential for an AI agent, existing literature has rarely explored the intersection of these fields in a systematic, comprehensive, and quantitative manner.


In Artificial Intelligence (AI), generalization refers to the model's robustness to out-of-distribution data related to the given task, beyond the data it was trained on. Existing solutions for out-of-domain generalization often rely on increasing data diversity \cite{li2020shape, hendrycks2019augmix, islam2021shape, doerig2023neuroconnectionist}, self-supervised learning \cite{thakral2023selfsupervised}, and meta-learning \cite{ho2022semi, NEURIPS2019_f4dd765c}. 

Concurrently, in the continual learning setting, agents are tasked with incrementally learning a sequence of tasks without experiencing catastrophic forgetting of the in-distribution data of earlier tasks. This challenge of catastrophic forgetting has led to the development of regularization-based approaches \eg, \cite{kirkpatrick2017overcoming,zenke2017continual,aljundi2018memory,li2017learning,castro2018end,wu2019large,douillard2020podnet,mittal2021essentials}. These methods aim to preserve essential parameters for old tasks through knowledge distillation or heuristics. Replay-based approaches, \eg, \cite{robins1995catastrophic,rebuffi2017icarl,castro2018end,wu2019large,hou2019learning,liu2020mnemonics},  involve storing or synthesizing exemplars from old tasks to train alongside new ones. More recently, architecture expansion methods, \eg, \cite{han2023enhancing}, have emerged, altering the model structure or tuning the learnable prompts to accommodate new tasks.

\begin{figure}[t!]
\centering
\includegraphics[width=0.95\columnwidth]
{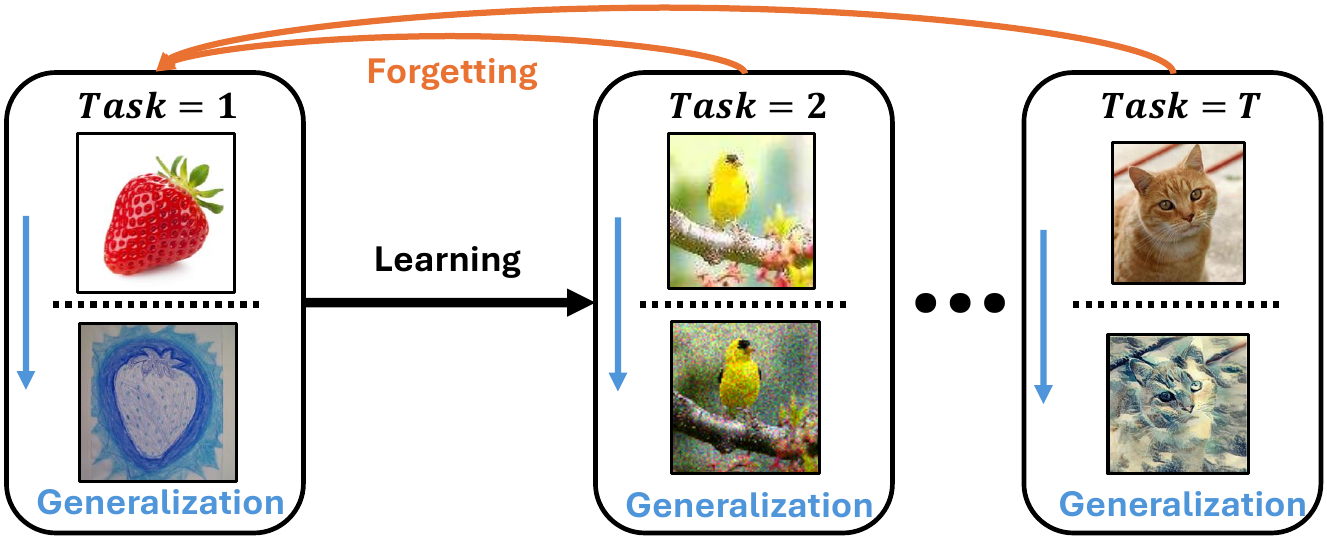}
\caption{\textbf{Illustration of the interplay between out-of-distribution generalization and continual learning.}
Over successive tasks, the model incrementally learns to recognize new object classes (black arrows), such as strawberries in Task 1 followed by birds in Task 2, and so on. During this continual learning process, the model exhibits a progressive loss of previously acquired knowledge known as catastrophic forgetting (orange arrows). Generalization refers to the model capability for performing well on the out-of-distribution data related to the given task, beyond the data it has seen during training (blue arrows). The sketch of the real-world strawberry in Task 1 refers to one out-of-distribution test sample, which was unseen by the model during training. Other examples for out-of-distribution data also include birds corrupted by noise or style-transferred cats. Here, we are to investigate the interplay of generalization within a given task and the forgetting about the previous tasks. 
}
\label{fig:fig1}
\vspace{-10pt}
\end{figure}

While several pioneering works apply generalization approaches for continual learning problems \cite{boschini2022transfer, riemer2018learning, peng2023lifelong, shi2023multi, ho2022semi},
none of these studies systematically explored the reciprocal effect between out-of-distribution generalization and forgetting within continual learning. In our initial efforts to close this research gap, we conduct a comprehensive study involving the direct integration of existing generalization methods with continual learning baselines, followed by an evaluation of these integrated models. We assess their performance in terms of forgetting about earlier tasks and handling out-of-distribution data from all the tasks trained so far (\textbf{Fig. \ref{fig:fig1}}). Remarkably, empirical evidence from our study indicates a mutually beneficial relationship between generalization and continual learning capabilities. Specifically, continual learning methods exhibiting reduced forgetting showcase enhanced generalization for all the previous tasks and the current task. Conversely, a superior generalization method contributes to a reduction in forgetting when it comes to earlier tasks. These findings underscore the intricate and reciprocal relationship between the two fundamental capabilities in AI.

Building upon this finding, we further enhance the existing generalization and continual learning baselines by introducing a simple yet effective Shape Texture Consistency Regularization method, dubbed STCR. In every current task, STCR distills both shape and texture representation of the training images by regularizing the logits of their style transferred versions and themselves. Extensive experiments validate that our approach can be seamlessly integrated with existing continual learning methods, including replay-free approaches. Its performance significantly surpasses any single continual learning baselines or their combined versions with any existing generalization methods. 
This highlights the effectiveness of our STCR in simultaneously improving both generalization and continual learning capabilities.

We summarize our key contributions:
\begin{itemize}
\item 

Building upon pioneering works applying generalization approaches in continual learning, our work fosters connections between both fields by exploring their reciprocal effects over a sequence of tasks. Through a series of systematic and comprehensive experiments involving 10 established generalization and continual learning baselines and 8 evaluation metrics, our empirical evidence reveals interesting insights that generalization and continual learning capabilities are mutually beneficial.

\item 
To enhance the generalization capabilities of existing continual learning baselines, we have developed a simple and effective shape-texture consistency regularization method (STCR). Our method distills both shape and texture representations from the training images of the current task, subsequently enhancing generalization and thereby mitigating forgetting.

\item 
Our STCR can be seamlessly integrated with both existing replay and replay-free continual learning methods. 
The experimental results demonstrate significant improvements in both generalization and continual learning performances, surpassing these existing continual learning methods in isolation or when combined with established generalization techniques. 

\end{itemize}

The structure of the subsequent sections is as follows. In \textbf{Sec. \ref{sec:relatedwork}}, we survey related works and highlight the differences between our work and the existing literature. In \textbf{Sec. \ref{sec:interplay}}, we introduced empirical evidence on the interplay of generalization and forgetting based on existing generalization and continual learning methods. Next, building upon this evidence, we introduce our proposed STCR method in  \textbf{Sec. \ref{sec:stcr}}. Subsequently, we elaborate the experimental setups (\textbf{Sec. \ref{sec:setup}}) and conduct extensive analysis on STCR (\textbf{Sec. \ref{exp}}).
Finally, we present our conclusions and discuss future work in \textbf{Sec. \ref{conlusionfuture}}.






\section{Related work}\label{sec:relatedwork}
\subsection{Continual learning}
\label{sec:cl}
Continuous learning involves acquiring knowledge in a sequential manner by training a single neural network on a series of tasks. During this process, only data from the current task is utilized for training. One grand challenge in continual learning is catastrophic forgetting \cite{mccloskey1989catastrophic,goodfellow2013empirical}.
To tackle this challenge, the following major types of approaches have been proposed.

Regularization-based approaches \cite{kirkpatrick2017overcoming,zenke2017continual,aljundi2018memory} impose stringent constraints on model parameters. This is achieved by penalizing changes in parameters that are crucial for retaining knowledge related to older tasks. 
Despite demonstrating some success in alleviating forgetting, these methods fall short of delivering satisfactory performance in demanding continual learning scenarios \cite{van2019three,wu2019large}. 
Recent approaches have incorporated knowledge distillation \cite{hinton2015distilling} as a regularization technique to minimize the changes to the decision boundaries of old tasks while learning new tasks. Methods based on knowledge distillation \cite{li2017learning,castro2018end,wu2019large,douillard2020podnet,mittal2021essentials}, typically involve maintaining a snapshot of the model trained on earlier tasks in a memory buffer. Subsequently, the knowledge encapsulated in the stored old model is distilled and transferred to the current model as part of the learning process. 
Recent approaches have adopted similar ideas of consistency regularization \cite{shi2023multi} on prototypes for class-incremental learning. 

Exemplar replay methods \cite{robins1995catastrophic,rebuffi2017icarl,castro2018end,wu2019large,hou2019learning,liu2020mnemonics} preserve a subset of representative examples from prior tasks in a memory buffer. Many of these approaches \cite{rebuffi2017icarl,castro2018end,wu2019large,hou2019learning} leverage herding heuristics \cite{welling2009herding} for selecting exemplars. The stored exemplars, in conjunction with new data, are then utilized to optimize the network parameters during the learning process of a new task. While replay strategies can be notably effective, they come with certain drawbacks. The replay of a restricted set of stored examples may result in overfitting. Additionally, storing a substantial number of images for replay purposes can be memory-intensive. 

To address memory constraints, \cite{shim2021online, sun2024cecr} proposed a scoring method to select what examples to store and \cite{sun2024cecr} introduced the memory update method to decide what examples in the memory buffer to replace. To avoid the problem of overfitting due to a limited number of replay samples, generative replay methods integrate data from new tasks with synthetic data generated by generative models, aiming to replicate stimuli encountered in previous instances \cite{shin2017continual,robins1995catastrophic,liu2020mnemonics,atkinson2018pseudo,liu2020generative,shen2021generative,van2021class}. Nevertheless, the generative models required for generating suitable synthetic data still tend to be sizable, memory-intensive, and challenging to train. 

To eliminate these issues, replay-free approaches have been proposed \cite{goswami2024fecam, petit2023fetril, zhu2021prototype}. These approaches often rely on augmenting prototypes \cite{zhu2021prototype}, generating pseudo features with prototypes as inputs \cite{goswami2024fecam}, and computing feature covariance relations for classification based on prototypes \cite{petit2023fetril}.

%

Dynamic architecture expansions \cite{aljundi2017expert,rajasegaran2019random,hung2019compacting,yan2021dynamically,hu2023dense} address the challenge of catastrophic forgetting by expanding the network. In this strategy, a new network is trained for each task, while the preceding networks are held constant. This ensures that the initially generated features for earlier task classes remain preserved. Despite showcasing enhanced performance compared to single-model approaches, these methods result in models that quickly escalate in size as the number of tasks increases. As a result, the scalability issue renders them impractical for many real-world applications.


Despite many of these advancements in continual learning, a predominant focus remains on testing models with in-distribution data. The generalization ability of these methods on out-of-distribution data was yet comprehensively evaluated. In a pioneering effort, we provide empirical evidence suggesting that continual learning methods demonstrating reduced forgetting of old tasks also exhibit improved generalization for all the trained tasks.

\subsection{Generalization}
\label{sec:shape-texture}
Generalization methods initially focused on increasing data diversity with augmentation techniques \cite{zhang2017mixup,devries2017improved,cubuk2018autoaugment, geirhos2018imagenet, hendrycks2019augmix, hermann2020origins}. 
Subsequently, researchers delved into adversarial training as a means to enhance generalization \cite{volpi2018generalizing,yi2021improved,lopes2019improving}. However, such adversarial training approaches often result in compromised performance within the training distribution itself \cite{tsipras2018robustness}.

An alternative avenue of research aimed at improving out-of-domain generalization involves exploring shape and texture representation learning. Geirhos et al. \cite{geirhos2018imagenet} uncovered that convolutional networks trained on natural images tend to acquire texture representations that generalize well to in-distribution data but exhibit sub-optimal performance on out-of-distribution data. In response to this, researchers have introduced methods focused on shape representation learning \cite{geirhos2018imagenet,hermann2020origins,shi2020informative}, often at the expense of in-distribution performance. Li et al. \cite{li2020shape} propose shape-texture debiased training using a mixup loss to prevent bias toward either shape or texture, striking a performance balance between in-distribution and out-of-distribution data.

In contrast to existing works\cite{ho2022semi, shi2023multi, riemer2018learning, boschini2022transfer, peng2023lifelong}, our research delves into the impact of generalization methods on forgetting over a sequence of tasks within the continual learning framework. 
While several pioneering works apply generalization approaches, such as data augmentation \cite{ho2022semi} with text label using BERT\cite{devlin2018bert}, transfer learning \cite{boschini2022transfer,peng2023lifelong}, and meta-learning \cite{riemer2018learning} for continual learning problems, none of these studies systematically explored the effect of out-of-distribution generalization in every task on catastrophic forgetting of earlier tasks. 
\cite{10444954} provided a comprehensive survey of continual learning theories, methods, and applications, highlighting the stability-plasticity trade-off and the need for strong intra- and inter-task generalization. Consistent with this framework, our empirical results emphasize the importance of enhancing intra-task generalization in continual learning. 
Building upon this evidence, we introduce a straightforward yet effective shape-texture consistency regularization approach tailored for continual learning. The experimental results demonstrate that our model, serving as a plug-and-play module seamlessly integrated with both replay and replay-free methods, outperforms various combinations of existing generalization and continual learning baselines by a substantial margin. We further validated the effectiveness of STCR through a loss landscape analysis similar to \cite{10444954}.


\section{interplay of generalization and forgetting}
\label{sec:interplay}

We started this section by formulating the problem setups (\textbf{Sec. \ref{subsec:problemsetting}}) followed by introducing competitive baselines, datasets, and protocols in continual learning (\textbf{Sec. \ref{subsec:baselines}}
).
We experimented on continual learning baselines and produced empirical evidence on the interplay of generalization and less forgetting in \textbf{Sec. \ref{empirical}}. Note that we use the common evaluation metrics, such as Forgetting $\mathcal{F}$ and corruption error $\mathcal{R}$-C, in the literature \cite{hendrycks2019benchmarking}. To quantify the bidirectional effects between generalization and forgetting, we also introduced $\Delta \mathcal{R}$-C and $\Delta \mathcal{F}$. See the detailed introduction to these metrics in \textbf{Sec. \ref{metrics}}. Moreover, to provide insights into the reasons why generalization eliminates forgetting, we presented loss landscape analysis in \textbf{Sec. \ref{subsec:lossland}}.  

\subsection{Problem setting}
\label{subsec:problemsetting}

We focus on the class-incremental scenario in continual learning, where the objective is to learn a unified classifier over incrementally occurring sets of classes \cite{rebuffi2017icarl}. Formally, let $\mathcal{T}=\{\mathcal{T}_1,...,\mathcal{T}_T\}$ be a sequence of classification tasks, $\mathcal{D}_t=\{(x_{i,t},y_{i,t})\}_{i=1}^{n_t}$ be the labeled training set of the task $\mathcal{T}_t$ with $n_t$ samples from $m_t$ distinct classes, and $(x_{i,t},y_{i,t})$ be the $i$-th (image, label) pair in the training set of the task $\mathcal{T}_t$. A single model $\Theta$ is trained to solve the tasks $\mathcal{T}$ incrementally. The model $\Theta$ consists of a feature extractor and a unified classifier. For the initial task $\mathcal{T}_1$, the model $\Theta_1$ learns a standard classifier for the first $m_1$ classes. In the incremental step $t$, only the classifier is extended by adding $m_t$ new output nodes to learn $m_t$ new classes, leading to a new model $\Theta_t$. During the training of $\Theta_t$, only the training set $\mathcal{D}_t$ for the task $\mathcal{T}_t$ is available. In the testing phase, $\Theta_t$ is required to classify all classes seen so far, \ie, $\{m_1,...,m_t\}$. Unlike the existing class-incremental works, $\Theta_t$ should perform well on both the in-distribution and out-of-distribution test data. For example, if $\Theta_t$ is trained to recognize real-world dog images, it should be capable of recognizing both real-world dog images and cartoon dog images during testing.

For replay-based continual learning methods, a memory buffer is often employed to store the original training images from the earlier tasks. We denote this memory buffer as the exemplar set $\mathcal{P}$. For fair comparisons, in all the experiments, we allow $\mathcal{P}$ to store 20 exemplars per object class by default, unless otherwise specified. The terms ``exemplar" and ``original training image" in the memory buffer are used interchangeably throughout the text.  

\subsection{Baselines and Protocols}
\label{subsec:baselines}

We introduce a list of competitive continual learning baselines used in this experiment:

\noindent \textbf{Naive} trains the model on the sequence of tasks without any measures to prevent forgetting.

\noindent \textbf{Replay.} Replay-based methods are the most effective group of continual learning methods. We implement the naive replay method based on  \cite{rebuffi2017icarl}, which stores the images from the previous tasks in a memory buffer and replays these images together with the training images in the current task. Consistent with \cite{rebuffi2017icarl}, we used a memory buffer size of 20 images per class.

\noindent \textbf{Gdumb} \cite{prabhu2020gdumb}  greedily stores upcoming samples while balancing the class distribution of the memory buffer. During testing, the model is trained from scratch only using the samples in the memory buffer.

\noindent \textbf{Lwf} \cite{li2017learning} is a knowledge distillation approach that regularizes the training process by aligning the predictions of the new model with those of the previous model on the new images at the current task. 

\noindent \textbf{iCARL} \cite{rebuffi2017icarl} 
combines Replay and a knowledge distillation technique introduced above in Lwf \cite{li2017learning}. During training, an exemplar set of images will be dynamically selected based on feature similarity with the prototypes. Image augmentation is also applied during training and replays.

\noindent \textbf{Cumulative} is an upper bound. It involves training the model on all the aggregated data from $D_{1}$ at $\mathcal{T}_1$ to $D_{t}$ at the current task $\mathcal{T}_t$, ensuring that the model never forgets about the old knowledge while learning the current task.

We benchmarked all these continual learning baselines on ImageNet-100 \cite{deng2009imagenet}. We used ImageNet-100-C \cite{hendrycks2019benchmarking} as its out-of-distribution generalization test set. All the models are trained on the first task $\mathcal{T}_1$ containing 50 classes, followed by 10 classes in every subsequent task. There are a total of 6 tasks. See \textbf{Sec. \ref{datasets}} and \textbf{Sec. \ref{protocols}} for details on datasets and protocols.

\subsection{Empirical evidence on the interplay of generalization and less forgetting reveal their mutual benefits}
\label{empirical}

\begin{figure}[t]
  \centering
  \includegraphics[width=0.9\columnwidth]{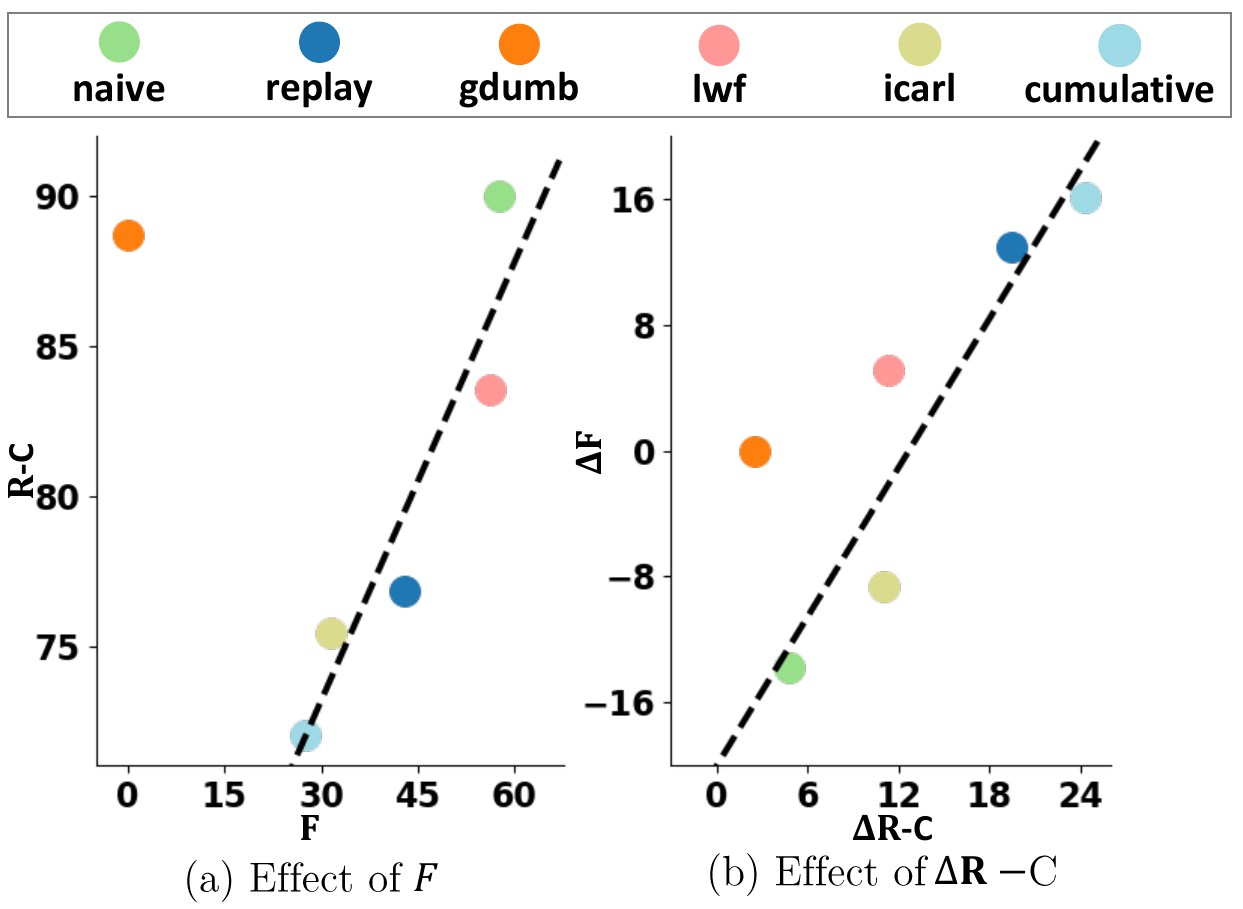}
  \caption{\textbf{Empirical evidence about the interplay of generalization and forgetting in continual learning on ImageNet-100 when $T=6$.}
  Subplot (a) illustrates the forgetting ($\mathcal{F}$) and generalization capability ($\mathcal{R}$-C) across a range of continual learning methods. For each continual learning algorithm, we report its $\mathcal{R}$-C and $\mathcal{F}$. 
  See \textbf{Sec. \ref{metrics}} for the definitions of $\mathcal{R}$-C and $\mathcal{F}$. We also performed a linear fitting (dashed line) between $\mathcal{R}$-C and $\mathcal{F}$ based on all the sample points on the subplot using RANSACRegressor \cite{teoh2015random}.  
  Subplot (b) presents the impact of generalization on reduced forgetting. We computed the performance difference in $\mathcal{R}$-C and $\mathcal{F}$ between existing continual learning baselines with and without the B-Aug generalization baseline, and denote these differences as $\Delta \mathcal{F}$ and $\Delta \mathcal{R}$-C. Postive $\Delta \mathcal{F}$ and $\Delta \mathcal{R}$-C imply that the continual learning model has improved generalization capability and reduced forgetting after integrating the B-Aug generalization algorithm, compared to its counterpart without B-Aug.
  The dashed line indicates the result of linear fitting between $\Delta F$ and $\Delta \mathcal{R}$-C using RANSACRegressor \cite{teoh2015random}.  
  See \textbf{Sec. \ref{subsec:baselines}} for the introduction of the continual learning baselines. See \textbf{Sec. \ref{a-baselines}} for the introduction of the generalization baselines. 
  See \textbf{Sec. \ref{metrics}} for the introduction of the evaluation metrics. }
  \label{fig:relationship}
  \vspace{-15pt}
\end{figure}

Here, we first examine how continual learning techniques, aimed at mitigating catastrophic forgetting, influence the overall generalization capabilities of models. Conversely, we delve into the impact of generalization methods on reducing forgetting across a sequence of tasks.
We report both results in \textbf{Fig. \ref{fig:relationship}}.

From \textbf{Fig. \ref{fig:relationship}a}, we note a strong positive linear correlation between $\mathcal{F}$ and $\mathcal{R}$-C ($r=0.93$ with a p-value of $0.01$) across diverse continual learning methods. See \textbf{Sec. \ref{metrics}} for the definitions of $\mathcal{F}$ and $\mathcal{R}$-C. This suggests that methods exhibiting reduced forgetting also exhibit enhanced generalization capability. Essentially, a continual learning approach with diminished forgetting retains more historical knowledge, consequently augmenting the diversity of acquired features. This retained knowledge, in turn, contributes to the model's improved ability to generalize when confronted with out-of-distribution data from all previously encountered tasks.

From \textbf{Fig. \ref{fig:relationship}b}, a strong positive linear correlation between $\Delta \mathcal{F}$ and $\Delta \mathcal{R}-C $ ($r=0.80$ with a p-value of $0.02$) was observed across various continual learning methods with and without the Basic Augmentation (B-Aug) generalization method introduced in \textbf{Sec. \ref{a-baselines}}. Briefly, B-Aug is a data augmentation technique involving color jittering, color drops, Gaussian noise, and Gaussian blur \cite{marr1980theory}.

The strong positive linear correlation suggests that better generalization capability contributes to diminishing forgetting for earlier tasks. For instance, the Replay method (dark blue), incorporating the B-Aug, achieves a reduction of $12.90$ in $\mathcal{F}$ and $19.46$ in $\mathcal{R}$-C compared to its isolated counterpart. Similarly, the cumulative baseline (cyan) also experiences a substantial decrease in $\mathcal{F}$ and $\mathcal{R}$-C when integrated with the B-Aug generalization method. One plausible explanation is that the generalization method empowers the model to capture more invariant and generic features in the current task, proving beneficial for both knowledge transfer to novel tasks and preserving learned knowledge from older tasks due to flatter loss minima \cite{geiping2023how}. In the subsequent section, we provide detailed analysis of these empirical results. 

However, we observe two exceptions in \textbf{Fig. \ref{fig:relationship}b}. Specifically, the integration of the B-Aug generalization method has adversely affected the continual learning and generalization performances of Naive and iCaRL. Naive serves as a lower bound without any mechanisms to mitigate catastrophic forgetting. Consequently, the inclusion of the B-Aug generalization method exacerbates overfitting, resulting in a performance decline. In the case of iCaRL \cite{rebuffi2017icarl}, which has its own built-in data augmentation method during training, the impact of the B-Aug generalization method is mitigated. These built-in data augmentations in iCaRL include random cropping, horizontal flipping, and rotation. 

\subsection{Loss landscape analysis provides insights into spurious feature learning across tasks}
\label{subsec:lossland}



\begin{figure*}[t!]
  \centering
  \includegraphics[width=0.85\textwidth]{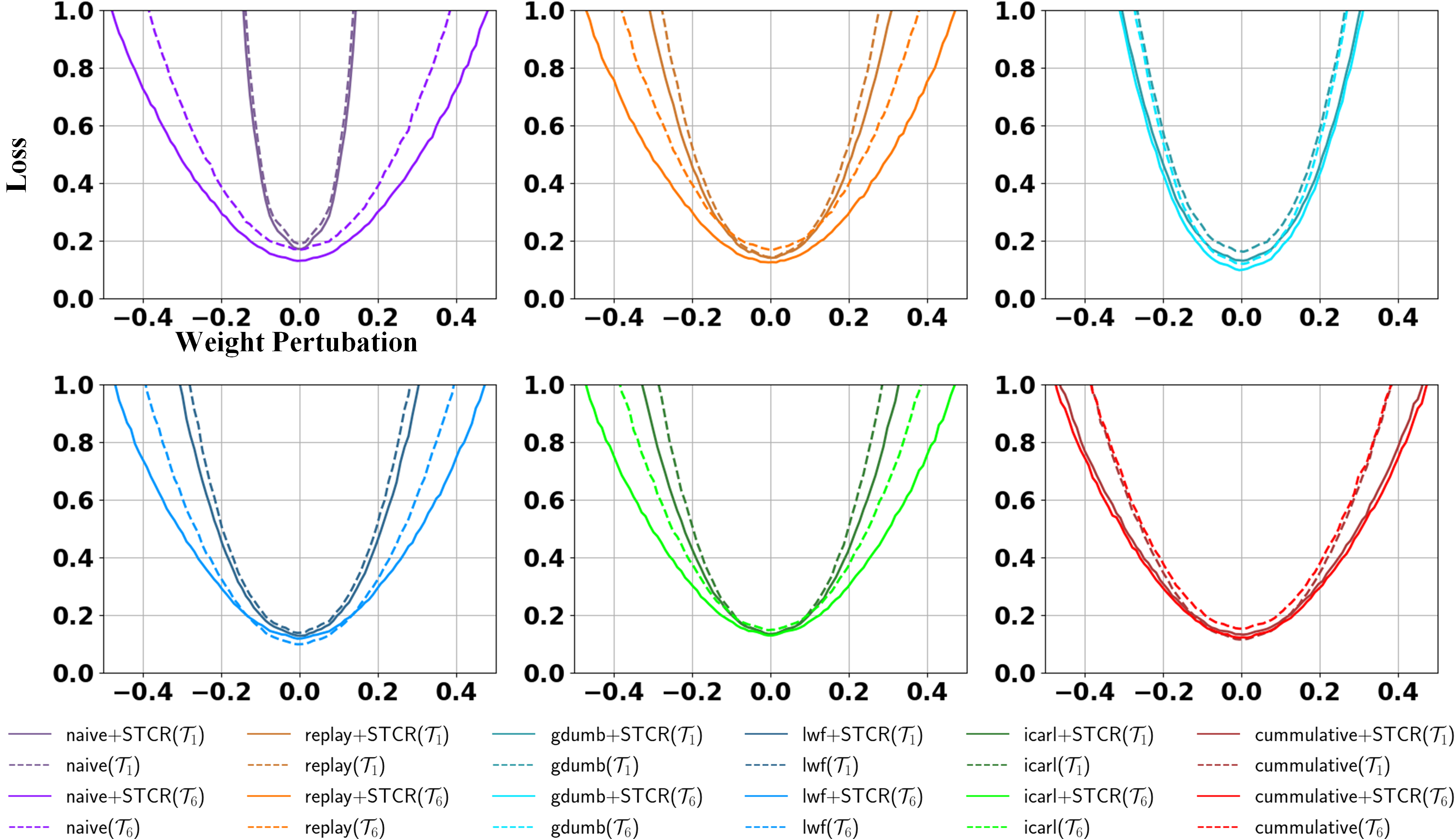}
  \caption{\textbf{Loss landscape analysis on existing continual learning methods with and without existing data augmentation techniques.} The loss landscape is produced by evaluating the final models at task $\mathcal{T}_6$ with perturbed weights in random directions on the test sets of tasks $\mathcal{T}_1$ (darker line) and $\mathcal{T}_6$ (lighter line) of the ImageNet-100 dataset. The x-axis indicates the magnitudes of weight perturbations. The results of the existing continual learning baselines in colors with the existing B-Aug generalization method (solid line) and without B-Aug (dotted line) are presented. See \textbf{Sec. \ref{subsec:baselines}} for the introduction of the continual learning baselines. See \textbf{Sec. \ref{a-baselines}} for the introduction of the generalization baselines. 
}
  \label{fig:landscape_existing}
  \vspace{-10pt}
\end{figure*}

In machine learning literature \cite{singla2021salient, izmailov2022feature, pmlr-v139-zhou21g, NEURIPS2022_fb64a552, 10.1145/3442188.3445883}, a spurious feature is an irrelevant pattern that a model can learn to capture. These features are correlated with the correct class labels in the training data but are not inherently relevant to the learning problem. A model with better generalization ability is less likely to capture spurious features. This encourages the model to make class predictions based on generic features, ensuring it is more robust to input perturbations. 

To quantify the extent to which models capture generic features for the current task, we computed the loss landscape \cite{li2018visualizing, fort2019deep} on the test set of $\mathcal{T}_6$ using the final models $\Theta_6$ trained after the last task $\mathcal{T}_6$. Specifically, to compute the loss landscape, we perturbed the weights of the models in random directions and tested the models on the current tasks. We reported their losses on 
$\mathcal{T}_6$ as a function of weight perturbation magnitudes in \textbf{Fig. \ref{fig:landscape_existing}}. From the results, we observed that the models with the B-Aug generalization method have wider loss landscapes compared to their counterparts in the same task. This implies that the generalization methods help the models learn to capture more generic features for object recognition in the current task.

In continual learning, spurious features are specific to individual tasks and do not generalize well. In other words, if a model captures spurious features for a particular task, these features are less effective for the other tasks. As a result, the models prone to capturing spurious features may struggle with earlier tasks, leading to increased forgetting. To validate this point, we performed another loss landscape analysis on the test set of the first task $\mathcal{T}_1$ using the final model $\Theta_6$ at $\mathcal{T}_6$.
As shown in \textbf{Fig. \ref{fig:landscape_existing}}, final models trained with B-Aug in $\mathcal{T}_6$ consistently have wider loss landscapes even on the test sets of $\mathcal{T}_1$, indicating better generalization. This suggests that these models capture more generic features shared across tasks, thereby reducing catastrophic forgetting in earlier tasks.


\section{Our proposed Method -- STCR}
\label{sec:stcr}
\begin{figure}[t!]
\centering
\includegraphics[width=0.9\columnwidth]{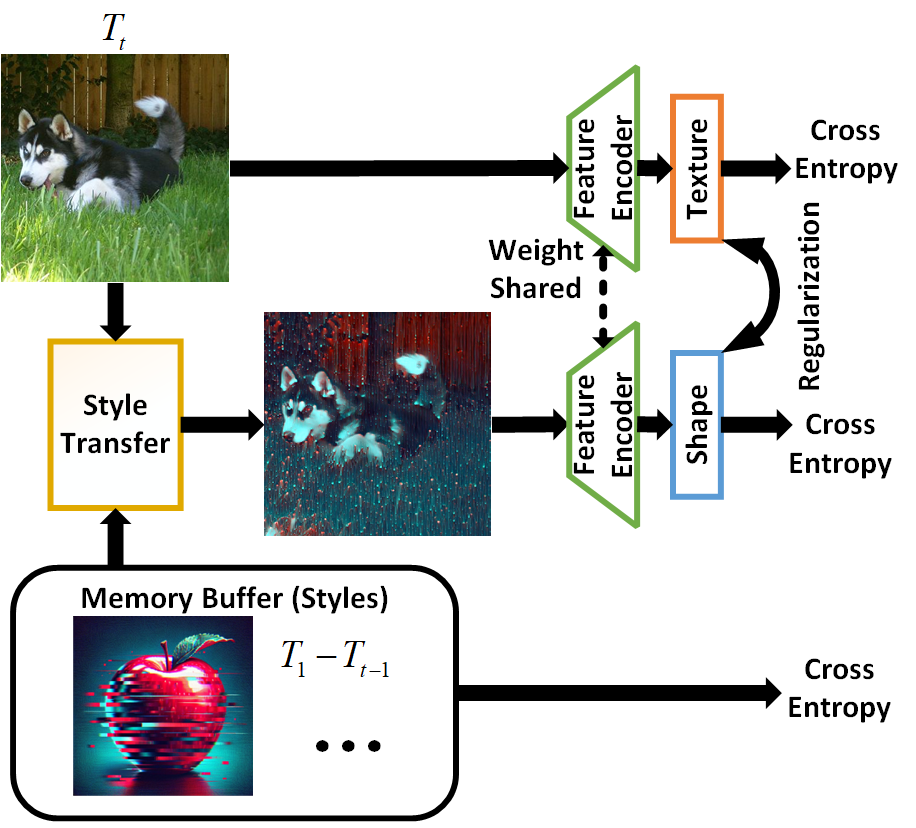}
\vspace{-4pt}
\caption{\textbf{Schematic of our shape-texture consistency regularization method (STCR).} Given an image from the training set, we create its shape-texture conflict counterpart by performing style transfers on the training images of the current task. The style images come from the examples stored in the memory buffer. The original image and shape-texture conflict image are combined in the mini-batch for training. The feature encoder extracts both texture-biased and shape-biased representations from these mini-batches. Their output logits are then normalized for classification via cross-entropy losses. Moreover, these logits are also regularized to make consistent class distributions. 
During replays, the model also rehearses the old images using cross-entropy losses. By using the proposed STCR, the resulting model can effectively prevent forgetting, and meanwhile, generalize to corrupted data, and domain-shifted data during inference. See \textbf{Sec. \ref{sec:stcr}} for the detailed design of our STCR.
}
\label{fig:fig2}
\vspace{-15pt}
\end{figure}

We present empirical evidence in \textbf{Sec. \ref{empirical}} highlighting how superior generalization performance in all the tasks trained so far contributes to reduced forgetting during continual learning. To enhance the generalization capability of established continual learning methods, we introduce a straightforward and effective regularization approach, dubbed Shape-Texture Consistency Regularization (STCR). See \textbf{Fig. \ref{fig:fig2}} for the method schematic. 
Our STCR can seamlessly integrate with any existing continual learning method. For instance, it can leverage exemplars in replay-based continual learning methods to enhance generalization ability. 
We introduce our STCR in this section.

\subsection{Shape-texture consistency regularization}
\label{sec:std}

Existing works in object recognition have suggested that models trained with original naturalistic images are often biased toward textures \cite{geirhos2018imagenet}. Augmenting the training data with style-transferred images such as shape-texture conflict images is an effective way to encourage models to learn shape-biased features, which are more generic for object recognition \cite{geirhos2018imagenet}. Inspired by these works, our STCR is designed to learn shape and texture representations for each classification task, enabling generalization to both in-distribution and out-of-distribution data. While standard training on natural images tends to learn texture representations for in-distribution generalization, shape representations are more robust to distribution shifts. To encourage the model to learn shape representations, we adopt the approach of Geirhos et al. \cite{geirhos2018imagenet} by using shape-texture conflict images for training. However, rather than use external artistic paintings in \cite{geirhos2018imagenet} for style transfers, we use different images within the current training set $\mathcal{D}_t$ as style images. This is because storing paintings in memory is not feasible for memory-constrained continual learning.

Specifically, for any generic continual learning methods with or without replay buffers, during optimization at the incremental step $t$, we randomly sample two mini-batches of images with a size of $k$ per batch from the current training set $\mathcal{D}t$. 
Next, we use one mini-batch of images as style templates and the other as content images. In this case, both mini-batches come from the current training set and they are randomly paired for style transfers. Every pair of style and content images could be from the same or different object classes. Mathematically, we denote the first mini-batch $\mathcal{X}=\{x_{1,t},\ldots,x_{k,t}\}$
as content images and the second mini-batch $\hat{\mathcal{X}}=\{\hat{x}_{1,t},\ldots,\hat{x}_{k,t}\}$ as style images. 
Together with these paired images, we can then generate a mini-batch of shape-texture conflict images $\tilde{\mathcal{X}}=\{\tilde{x}_{1,t},\ldots,\tilde{x}_{1,t}\}$ based on $\mathcal{X}$ and $\hat{\mathcal{X}}$,
where
\begin{equation}
    \tilde{x}_{i,t} = f\big(x_{i,t},\hat{x}_{i,t}\big).
\label{eq:style}
\end{equation}
Here $f(\cdot)$ is the style transfer operation. 
Specifically, we use the real-time style transfer approach, AdaIN \cite{huang2017arbitrary}, pre-trained using MS-COCO \cite{lin2014microsoft} 
and WikiArt \cite{WikiArt}. 
Note that none of these datasets overlap with the datasets we use for continual learning (see \textbf{Sec. \ref{datasets}} for the details about our datasets). To examine the quality of style transferred images by AdaIN which AdaIN has never been trained on, we provide additional visualization results in the Appendix (\ref{appendix:styles}) and verify that the quality of those generated images is reasonable.  Moreover, we emphasize that AdaIN \cite{huang2017arbitrary} is used as a proof-of-concept for style transfers. There are many other style transfer methods available, such as \cite{zhang2018multi, sheng2018avatar, Reich2021}, which are more compute-efficient for real-world applications.


When training solely on the original natural images, the model tends to acquire texture representations. Conversely, training exclusively on shape-texture conflict images leads to the learning of shape representations. To concurrently develop both shape and texture representations, we utilize a consistency regularization approach that involves both the original natural images and the shape-texture conflict images. Specifically, let $\mathcal{X}=\{x_{1,t},\ldots,x_{k,t}\}$ be a mini-batch of $k$ images randomly sampled from the current training set $\mathcal{D}_t$ in the incremental step $t$, and $\tilde{\mathcal{X}}=\{\tilde{x}_{1,t},\ldots,\tilde{x}_{k,t}\}$ be the corresponding mini-batch of shape-texture conflict images generated following \textbf{Eq. (\ref{eq:style})}. Next, we obtain a new mini-batch $\{\mathcal{X},\tilde{\mathcal{X}}\}$ by combining $\mathcal{X}$ and $\tilde{\mathcal{X}}$ for model training. Let $\mathcal{Z}=\{z_{1,t},\ldots,z_{k,t}\}$ and $\tilde{\mathcal{Z}}=\{\tilde{z}_{1,t},\ldots,\tilde{z}_{k,t}\}$ be the corresponding logits of $\mathcal{X}$ and $\tilde{\mathcal{X}}$, produced by the model $\Theta_t$, respectively. $\mathcal{Z}$ is generated using original natural images as inputs and encodes texture information. Conversely, $\tilde{\mathcal{Z}}$ is generated using shape-texture conflict images as input and encodes shape information. To ensure that the model captures both texture and shape information, we encourage $\mathcal{Z}$ and $\tilde{\mathcal{Z}}$ to predict similar class distributions with a consistency regularization loss:
\begin{equation}
    \mathcal{L}^{STCR} = D_{KL}(p||q)+D_{KL} (q||p),
\label{eq:shape-texture}
\end{equation}
where $p=\sigma(\mathcal{Z}/\tau)$, $q=\sigma(\tilde{\mathcal{Z}}/\tau)$,
$\tau$ is a temperature hyper-parameter controlling the smoothness of the probability distributions. We empirically set it to $2$.
$D_{KL}$ is the Kullback-Leibler divergence loss comparing the distance between two distributions. $\sigma(\cdot)$ is softmax function. Our consistency regularization loss only requires a mini-batch of training images from the current task and it can be computed in an online manner.

\subsection{Refinement of STCR for replay-based continual learning}
\label{continualSTCR}

Just like many existing generalization methods, $\mathcal{L}^{STCR}$ can be integrated with any established continual learning baselines, such as
EWC \cite{kirkpatrick2017overcoming}, exemplar replay \cite{rebuffi2017icarl}, or knowledge distillation \cite{li2017learning}. 
The previous subsection (\textbf{Sec. \ref{sec:std}}) describes the generic STCR, where both style and content images come from the current task's training set $\mathcal{D}_t$. However, replay-based continual learning methods use memory buffers $\mathcal{P}$ to store exemplars. We can leverage these exemplars to refine our STCR for replay-based methods, as detailed in this subsection.

In replay-based continual learning methods, training images from earlier tasks are often stored in the memory buffer $\mathcal{P}$ for replays in the current task. These stored images often retain additional information from earlier tasks. We capitalized on these exemplars $\mathcal{P}$ to enhance our STCR. Specifically, instead of using training images as style templates from the current task, we employed images from the memory buffer $\mathcal{P}$ as style templates and kept the training images from the current task as content to generate shape-texture conflict images. These exemplar sets $\mathcal{P}$ in the memory buffer encompass a broader range of classes from previous tasks, enhancing the diversity of the shape-texture conflict images. 
Simultaneously, these conflict images carry styles from earlier tasks, serving as a form of feature rehearsal to mitigate forgetting. In our experiments, we also observed that applying shape-texture consistency regularization during replays introduces noise from shape-texture conflict images, which can detrimentally impact replay performances (refer to \textbf{Sec. \ref{sec:discussion}}). 

The schematic of our STCR is illustrated in \textbf{Fig.} \ref{fig:fig2}, and its implementation is outlined in \textbf{Algorithm} \ref{algorithm}. 
At every incremental step $t$, the overall loss function of STCR to train a neural network is defined as:
\begin{equation}
   \mathcal{L}_t = \mathcal{L}_{\mathcal{D}_t,\tilde{\mathcal{D}}_t}^{CE}
   +\beta\mathcal{L}_{\mathcal{P}}^{CE}+\gamma \mathcal{L}_{\mathcal{D}_t,\tilde{\mathcal{D}}_t}^{STCR},
\label{eq:loss}
\end{equation}
where cross-entropy loss $\mathcal{L}_{\mathcal{D}_t,\tilde{\mathcal{D}}_t}^{CE}$ is computed on the combination of shape-texture conflict images $\tilde{\mathcal{D}}_t$ and the original natural images $\mathcal{D}_t$ from the current task $t$. For the replay-based continual learning methods, $\mathcal{L}_{\mathcal{P}}^{CE}$ is computed only on the original natural images from exemplar sets $\mathcal{P}$. 
The hyperparameter $\beta = 1$ and $\gamma = 0.01$ regulate the loss for exemplar replays and the shape-texture consistency regularization loss respectively. We analyzed the effect of $\beta$ and $\gamma$ in \textbf{Sec.\ref{sec:discussion}}.

\subsection{Implementation details.} 
The experiments are conducted based on the publicly available official code provided by Mittal \etal~\cite{mittal2021essentials}. For ImageNet-100 and ImageNet-1000, we use an 18-layer ResNet with randomly initialized weights. The network is trained for 70 epochs in the initial base task with a base learning rate of 1e-1, and is trained for 40 epochs in the incremental task with a base learning rate of 1e-2. The base learning rate is divided by 10 at epochs $\{30,60\}$ in the initial base task, and is divided by 10 at epochs $\{25,35\}$ in the incremental steps. For CIFAR-100, we use a 32-layer ResNet with randomly initialized weights. The network is trained for 120 epochs in the initial base task with a base learning rate of 1e-1, and is trained for 60 epochs in the subsequent incremental tasks with a base learning rate of 1e-2. We adopt a cosine learning rate schedule, in which the learning rate decays until 1e-4. For all networks, the last layer is cosine normalized, as suggested in \cite{hou2019learning}. All networks are optimized using the SGD optimizer with a mini-batch of 128, a momentum of $0.9$, and a weight decay of 1e-4. Following \cite{hou2019learning,mittal2021essentials}, we use an adaptive weighting function for knowledge distillation loss. At each incremental step $t$, $\lambda_t=\lambda_{base}\big (\sum_{i=1}^t {m_i}/m_t\big)^{2/3}$ where $\sum_{i=1}^t {m_i}$ denotes the number of all seen classes from step $1$ to step $t$, and $m_t$ denotes the number of classes in step $t$. $\lambda_{base}$ is set to $20$ for CIFAR-100, $100$ for ImageNet-100, and $600$ for ImageNet-1000, as suggested by \cite{mittal2021essentials}. 
To ensure statistical significance, we conducted 3 runs for all the experiments and reported their means and standard deviations.
All codes will be made publicly available upon publication.

\section{Experimental setup}
\label{sec:setup}


\subsection{Datasets}
\label{datasets}
Same as previous continual learning works \cite{hou2019learning,liu2020mnemonics,tao2020topology,douillard2022dytox}, we perform class-incremental experiments on three standard image datasets CIFAR-100 \cite{krizhevsky2009learning}, ImageNet-100, and ImageNet-1000 \cite{deng2009imagenet} and follow the same training and test data splits.
CIFAR-100 contains $60,000$ color images with the image size of $32 \times 32$ from 100 classes, in which $50,000$ images are for training and the remaining $10,000$ are for testing. 
ImageNet-1000 contains around $1.3$ million color images of size $224 \times 224$ from $1,000$ classes. For each class, there are $1,300$ images for training and $50$ images for validation. 
ImageNet-100 is a subset of Imagenet-1000 with $100$ classes that are randomly sampled using the NumPy seed of $1993$. 

Since there is a lack of out-of-distribution test sets in the existing continual learning literature, we use the datasets ImageNet-1000-C \cite{hendrycks2019benchmarking}, CIFAR-100-C \cite{hendrycks2019benchmarking}, and ImageNet-1000-R \cite{hendrycks2021many} for evaluating out-of-distribution generalization. ImageNet-1000-C and CIFAR-100-C consist of $15$ diverse corruption types applied to validation images of ImageNet-1000 and CIFAR-100, respectively, with five different severity levels for each corruption type. ImageNet-1000-R is a real-world distribution shift dataset that includes changes in image styles, geographic locations, etc. 
We curated ImageNet-100-C and ImageNet-100-R from ImageNet-1000-C and ImageNet-1000-R 
also by random sampling with seed $1993$.

\subsection{Protocols} 
\label{protocols}
We follow the standard protocols used in \cite{hou2019learning,liu2020mnemonics,tao2020topology,douillard2020podnet,mittal2021essentials,douillard2022dytox} for benchmarking class-incremental learning methods in continual learning. The protocol consists of an initial base task $\mathcal{T}_1$ followed by $T$ incremental tasks, where $T$ is set to $6$ or $11$. For all datasets, we assign half of the classes to the initial base task and distribute the remaining classes equally over the incremental steps. 

\subsection{Metrics} 
\label{metrics}
We use the following metrics to measure the continual learning and generalization performance. Let $A_{j,i}$ denote the accuracy on task $i$ after training the continual learning model $\phi_j$ on task $j$. 

  \noindent \textbf{Average incremental accuracy (Acc.)} \cite{rebuffi2017icarl} is computed by averaging the accuracies of all the models $\{ \Theta_1, ..., \Theta_T\}$ obtained in all the incremental steps. Each model's accuracy is evaluated on all classes seen thus far. The mathematical formulation is listed below:
  \begin{equation}
    \text{Acc.} = \frac{1}{T} \sum_{t=1}^{T} \left( \frac{1}{t} \sum_{i=1}^{t}  A_{t,i} \right)
    \label{eq:avgac}
  \end{equation}
  
  Acc. is a common evaluation metric in the continual learning literature \cite{rebuffi2017icarl, mittal2021essentials, douillard2020podnet, kang2022class}, measuring the overall model performance on the in-distribution data over all the trained tasks. The higher the Acc., the better. 

\noindent  \textbf{Average Accuracy $A_T$}
computes the mean accuracy over all the test sets of all tasks for the final model $\phi_T$ trained after the last task $T$ in a continual learning protocol. We denote the average accuracy as $A_T$:
 \begin{equation}
      A_T = \frac{1}{T} \sum_{t=1}^{T}  A_{T,t}  
      \label{eq:act}
  \end{equation}

  \noindent \textbf{Forgetting ($\mathcal{F}$)} \cite{liu2020mnemonics} measures the performance drop of $\Theta_T$ on the initial task $\mathcal{T}_1$. It is the gap between the accuracy of $\Theta_1$ and the accuracy of $\Theta_T$ on the same in-distribution data of $\mathcal{T}_1$. Formally, $\mathcal{F}$ is defined as:
  \begin{equation}
    \mathcal{F} = A_{T,1} - A_{1,1}
    \label{eq:ac}
  \end{equation}
  In the continual learning literature, $\mathcal{F}$ reflects how much the model has forgotten about $\mathcal{T}_1$. The smaller $\mathcal{F}$, the better.

  \noindent \textbf{Generalization (\textbf{$\mathcal{R}$-C} and \textbf{$\mathcal{R}$-R}}). 
  Here, we introduce two evaluation metrics to assess the out-of-distribution generalization capabilities of a model. This includes the robustness against data corruption and the domain shifts. Consistent with \cite{hendrycks2019benchmarking}, we assess the robustness against data corruptions of the ultimate model $\Theta_T$ at the last task $T$ (\textbf{$\mathcal{R}$-C}) by calculating its mean corruption error across all the classes of ImageNet-100-C, ImageNet-1000-C, and CIFAR-100-C respectively. The \textbf{$\mathcal{R}$-C} of $\Theta_T$ can then be defined as its own mean corruption error normalized by the mean corruption error of AlexNet \cite{krizhevsky2012imagenet} ($mCE_{AlexNet}$):

  \begin{equation}
    \mathcal{R}\text{-C} = \frac{mCE_{model}}{mCE_{AlexNet}}
    \label{eq:rc}
  \end{equation}

Note that AlexNet is trained on the entire dataset in an offline setting without class incremental learning. Basic Augmentation (B-Aug) in \textbf{Sec. \ref{a-baselines}} is applied during the training of AlexNet.
  
  
  Following the method described in \cite{hendrycks2021many}, we also evaluate the generalization of the final model $\Theta_T$ against domain shifts (\textbf{$\mathcal{R}$-R}) with accuracy measurements on all the classes of ImageNet-100-R and ImageNet-1000-R. 
  
  

\noindent \textbf{Change in Forgetting and Robustness ($\Delta \mathcal{F}$ and $\Delta \mathcal{R}$-C)}. 
 To quantitatively assess the impact of generalization methods on reducing forgetting across a sequence of tasks, we introduce two evaluation metrics $\Delta \mathcal{F}$ and $\Delta \mathcal{R}$-C. 
Let $\mathcal{F}$ denote the forgetting score for the final model $\Theta_T$ trained using a continual learning method, with and without Basic Augmentation (B-Aug) as specified in \textbf{Sec. \ref{a-baselines}}. These forgetting scores are represented as $\mathcal{F}_{wB, \Theta_T}$ and $\mathcal{F}_{woB, \Theta_T}$ respectively. The difference in $\mathcal{F}$ can be defined as:
\begin{equation}
    \Delta \mathcal{F} = \mathcal{F}_{wB, \Theta_T} - \mathcal{F}_{woB, \Theta_T}
\label{eq:deltaF}
\end{equation}
Similarly, we can compute the difference in $\mathcal{R}$-C as $\Delta \mathcal{R}$-C between the continual learning baseline with and without B-Aug generalization algorithm:
\begin{equation}
    \Delta \mathcal{R}\text{-C} = \mathcal{R}\text{-C}_{wB, \Theta_T} - \mathcal{R}\text{-C}_{woB, \Theta_T}
\label{eq:deltaRC}
\end{equation}


\textbf{Backward Transfer (BWT)}. Backward Transfer (BWT) \cite{lin2022beyond, diaz2018don} measures the influence of learning a new task on the performance of previously learned tasks. A positive BWT indicates that learning a new task improves performance on earlier tasks, while a negative BWT suggests a performance decline due to catastrophic forgetting.

To compute BWT, we evaluate the model's accuracy on all previously learned tasks before and after training on a new task. Formally, BWT is defined as:

\begin{equation}
B W T=\frac{\sum_{i=2}^T \sum_{j=1}^{i-1}\left(A_{i, j}-A_{j, j}\right)}{\frac{T(T-1)}{2}}
\label{eq:bwt}
\end{equation}
  where $A_{j,i}$ denotes the accuracy on task $i$ after training on the $j$-th task.

Without loss of generality, we reported the results of Average Accuracy $A_T$ and Backward Transfer (BWT) 
in Appendix (\ref{appendix:metrics}). The conclusions are consistent with those obtained from other evaluation metrics.

\subsection{Generic continual learning frameworks}
\label{subsec:wkbaselines}

First, to demonstrate that our STCR can be integrated with any generic continual learning frameworks, we provide an overview of major frameworks used in continual learning (see \textbf{Sec.\ref{integrated}}) and combine our STCR with these frameworks. In \textbf{Sec. \ref{subsec:baselines}}, we introduced \textbf{Naive} and \textbf{Replay}. We list the other major frameworks below:


\noindent \textbf{Weight Regularization-based (WReg).} WReg-based methods are one category of continual learning methods. Elastic Weight Consolidation (EWC) \cite{kirkpatrick2017overcoming} is one of these methods. We implement EWC, which leverages Fisher Information matrix to identify important parameters and penalizes further changes of these parameters in the later tasks. 


\noindent \textbf{Knowledge distillation-based (KD).} KD methods are special cases of weight-regularization methods. They turn out to be very effective, especially when they are combined with a replay-based method. We implement the knowledge distillation loss from \cite{li2017learning}, which saves a snapshot of the model in the previous tasks and distills knowledge from the old model to the new model based on the training images at the current task. 

\subsection{Existing continual learning baselines}
\label{subsec:sotabaselines}

Next, to demonstrate that our STCR can further enhance the state-of-the-art (SOTA) continual learning algorithms, 
we integrate our STCR with three SOTA continual learning methods, each of which could incorporate a combination of multiple continual learning frameworks introduced above. We implement these approaches using their publicly available code.


\noindent \textbf{CCIL} \cite{mittal2021essentials} employs a hybrid approach combining replay and regularization strategies. 

\noindent \textbf{Podnet} \cite{douillard2020podnet} utilizes an efficient distillation loss across multiple spatial dimensions of feature maps after various pooling operations.

\noindent \textbf{AFC} \cite{kang2022class} estimates the relationship between the representation changes and the resulting loss increases incurred by model updates. The greater the loss increase on the old representations, the more important the weights are to the old task. Based on the importance of the weights, the model restricts updates of important features while allowing changes in less critical features, thereby preventing forgetting.

\subsection{Generalization baselines}
\label{a-baselines}

We compare our STCR against the following generalization approaches summarized below:

\noindent \textbf{Lower bound (LB)} does not use any techniques to encourage generalization. 

\noindent \textbf{Basic augmentation (B-Aug)} follows \cite{hermann2020origins} and applies
simple and naturalistic augmentations, including color distortion, noise, and blur. Specifically, we employ color jitter with an 80\% probability, color drop with a 20\% probability, Gaussian noise (mean=0 and std=0.025) with a 50\% probability, and Gaussian blur (kernel size is 10\% of the image width/height) with a 50\% probability.

\noindent \textbf{Auto-augmentation (A-Aug)} \cite{cubuk2018autoaugment} represents an automated technique for identifying data augmentation strategies from the dataset, frequently resulting in improved generalization compared to basic augmentation techniques.


\noindent \textbf{Style augmentation (S-Aug)} \cite{foret2020sharpness}
adopts \cite{geirhos2018imagenet} to create shape-texture conflict images using artistic painting styles and trains models on both shape-texture conflict images and original natural images. In contrast, our STCR method eliminates the need for external memory to store artistic-style images, and we introduce a regularization loss between original and shape-texture conflict images.


\noindent \textbf{Mixup loss (Mixup)} initially generates shape-texture conflict images by following \cite{zhang2017mixup} and then applies a mixup loss to these conflict images. The hyperparameter 
$\epsilon$, determining the relative importance of shape and texture, is set to 0.5, consistent with \cite{li2020shape}. 

It is important to note that our experimentation involves 4 generic continual learning frameworks, 7 continual learning baselines, and 5 generalization methods. Ideally, applying each generalization method to every continual learning baseline or framework would result in a total of $(4+7)\times 5 = 55$ combinations. However, due to limitations in computing resources, we pragmatically select a subset of continual learning methods and generalization methods for controlled experiments. In these experiments, we either vary the continual learning methods or vary the generalization methods, but not both simultaneously. Refer to \textbf{Sec. \ref{exp}} for specifications of each experiment.

\section{Experiments and Results}
\label{exp}

\subsection{STCL can be seamlessly integrated with any generic continual learning frameworks}
\label{integrated}


\begin{table}[t!]
\centering
\footnotesize
\resizebox{0.95\columnwidth}{!}{
\begin{tabular}{@{}lcccclcccc@{}}
    \toprule
    \multirow{2}{*}{Method} & \multicolumn{3}{c}{$T$=6} &  & \multicolumn{3}{c}{$T$=11} \\
    \cmidrule(lr){2-4} \cmidrule(lr){6-8}
     & $\mathcal{R}$-C$\downarrow$ & Acc. $\uparrow$ & $\mathcal{F}$ $\downarrow$ &  & $\mathcal{R}$-C$\downarrow$ & Acc. $\uparrow$ & $\mathcal{F}$ $\downarrow$ \\
    \midrule
    Naive & \makecell{113.49 \\ \scriptsize{(4.04)}} & \makecell{23.54 \\ \scriptsize{(0.27)}} & \makecell{81.81 \\ \scriptsize{(1.53)}} &  
    & \makecell{123.49 \\ \scriptsize{(5.54)}} & \makecell{11.55 \\ \scriptsize{(0.94)}} & \makecell{80.64 \\ \scriptsize{(0.93)}} \\
    \rowcolor{Gray} 
    Naive + STCR & \makecell{\textbf{105.95} \\ \scriptsize{(5.90)}} & \makecell{\textbf{30.65} \\ \scriptsize{(3.52)}} & \makecell{\textbf{75.65} \\ \scriptsize{(2.05)}} & 
     & \makecell{\textbf{110.91} \\ \scriptsize{(3.33)}} & \makecell{\textbf{24.15} \\ \scriptsize{(1.82)}} & \makecell{\textbf{78.01} \\ \scriptsize{(4.85)}} \\
    \midrule
    WReg  & \makecell{106.14 \\ \scriptsize{(7.59)}} & \makecell{54.56 \\ \scriptsize{(3.06)}} & \makecell{52.74 \\ \scriptsize{(3.51)}} & 
    & \makecell{112.42 \\ \scriptsize{(2.46)}} & \makecell{50.33 \\ \scriptsize{(4.63)}} & \makecell{56.37 \\ \scriptsize{(5.03)}} \\
    \rowcolor{Gray} 
    WReg + STCR & \makecell{\textbf{99.86} \\ \scriptsize{(6.67)}} & \makecell{\textbf{58.05} \\ \scriptsize{(1.25)}} & \makecell{\textbf{40.83} \\ \scriptsize{(3.35)}} &  
    & \makecell{\textbf{107.08} \\ \scriptsize{(1.75)}} & \makecell{\textbf{51.33} \\ \scriptsize{(2.71)}} & \makecell{\textbf{46.13} \\ \scriptsize{(3.46)}} \\
    \midrule
    Replay  & \makecell{96.59 \\ \scriptsize{(2.44)}} & \makecell{67.57 \\ \scriptsize{(0.45)}} & \makecell{31.30 \\ \scriptsize{(1.99)}} &  
    & \makecell{104.72 \\ \scriptsize{(8.15)}} & \makecell{52.56 \\ \scriptsize{(1.42)}} & \makecell{34.74 \\ \scriptsize{(6.48)}} \\
    \rowcolor{Gray} 
    Replay + STCR & \makecell{\textbf{84.86} \\ \scriptsize{(1.15)}} & \makecell{\textbf{69.10} \\ \scriptsize{(0.89)}} & \makecell{\textbf{24.87} \\ \scriptsize{(2.48)}} &  
    & \makecell{\textbf{93.30} \\ \scriptsize{(0.14)}} & \makecell{\textbf{62.34} \\ \scriptsize{(5.22)}} & \makecell{\textbf{28.31} \\ \scriptsize{(6.31)}} \\
    \midrule
    KD  & \makecell{91.23 \\ \scriptsize{(6.48)}} & \makecell{67.54 \\ \scriptsize{(1.73)}} & \makecell{11.76 \\ \scriptsize{(0.51)}} &  
    & \makecell{101.76 \\ \scriptsize{(10.09)}} & \makecell{55.01 \\ \scriptsize{(1.43)}} & \makecell{24.83 \\ \scriptsize{(2.61)}} \\
    \rowcolor{Gray} 
    KD + STCR & \makecell{\textbf{80.46} \\ \scriptsize{(3.12)}} & \makecell{\textbf{70.63} \\ \scriptsize{(2.07)}} & \makecell{\textbf{9.79} \\ \scriptsize{(0.95)}} &  
    & \makecell{\textbf{92.68} \\ \scriptsize{(2.36)}} & \makecell{\textbf{60.31} \\ \scriptsize{(2.18)}} & \makecell{\textbf{14.95} \\ \scriptsize{(0.15)}} \\
    \bottomrule
  \end{tabular}
}
\caption{
\textbf{Generalization and continual learning performance of the generic continual learning frameworks with and without our STCR.} We reported the performance of four generic continual learning frameworks with (highlighted in gray background) and without our STCR on ImageNet-100 with $T$=$6$ (first column) and $T$=$11$ (second column) in terms of their generalization ability against data corruption ($\mathcal{R}$-C), and continual learning ability (Acc., and $\mathcal{F}$). See \textbf{Sec. \ref{subsec:wkbaselines}} and \textbf{Sec. \ref{subsec:baselines}} for the introduction of these generic continual learning frameworks. See \textbf{Sec. \ref{metrics}} for the introduction of the evaluation metrics. The best results are in bold. The numbers in brackets indicate the standard deviation after 3 runs. 
}
\label{tab:benefit}
\vspace{-10pt}
\end{table}

Given the mutually beneficial relationship between generalization and reduced forgetting discussed in \textbf{Sec. \ref{empirical}}, we introduced STCR as a generalization method specifically designed for continual learning (\textbf{Sec. \ref{sec:stcr}}). Here, we demonstrate that our STCR can be seamlessly integrated with any generic continual learning framework. Specifically, we selected the three most commonly used continual learning frameworks as well as one naive framework (\textbf{Sec. \ref{subsec:baselines}}) and then applied STCR to them.
These integrated framewoks are labelled as ``the name of the framework + STCR". We evaluated ``framework + STCR"
on ImageNet-100 with two task sequence lengths of $T=6$ and $T=11$ (\textbf{Sec. \ref{tab:discussion}}). For comparisons, we also evaluate these continual learning frameworks without our STCR in the same experiment settings. We report their generalization and continual learning performances in \textbf{Tab. \ref{tab:benefit}}. 

From \textbf{Tab. \ref{tab:benefit}}, across both task sequence lengths $T=6$ and $T=11$, we observed that all the continual learning frameworks integrated with our STCR outperform their counterparts without STCR by a large margin, in terms of both generalization ($\mathcal{R}$-C) and continual learning performances (Acc. and $\mathcal{F}$). For instance, when $T=6$, WReg + STCR beats WReg alone by 6.28\% in $\mathcal{R}$-C, 3.49\% in Acc. and 11.91\% in $\mathcal{F}$. This implies that our STCR, when integrated with continual learning frameworks, is effective in enhancing generalization and reducing forgetting. 

Notably, among the four continual learning frameworks, Naive, WReg, and KD are replay-free frameworks and they do not require exemplar sets $\mathcal{P}$. Incorporating the generic STCR introduced in \textbf{Sec. \ref{sec:std}} into these frameworks also significantly reduces forgetting and enhances generalization. In particular, the naive continual learning framework, which lacks measures to prevent forgetting, often has chance-level continual learning performance. With STCR, we observed a dramatic boost in continual learning and generalization performance across all evaluation metrics (compare Naive + STCR versus STCR).  



Moreover, we also found that the positive effect of STCR becomes more prominent when it is integrated with those frameworks that are better continual learners. For example, the existing works have shown that KD and Replay are the most effective strategies in reducing catastrophic forgetting, which most of the SOTA continual learning methods adopt. Indeed, compared with Naive and WReg, this is also demonstrated by higher Acc. and lower $\mathcal{F}$ of KD and Replay as shown in \textbf{Tab. \ref{tab:benefit}}.
With our STCR, the generalization performance of Replay+STCR and KD+STCR gets boosted by 11.73\% and 10.77\% compared to their counterparts respectively. This is in high contrast with a relative improvement of 7.54\% and 6.28\% in $\mathcal{R}$-C for Naive+STCR and WReg+STCR.

\subsection{State-of-the-art continual learning methods with STCR beat their standalone configurations}

In the subsection above, we demonstrated the feasibility of integrating our STCR with any generic continual learning framework. Here, we specifically assess the impact of STCR on state-of-the-art (SOTA) continual learning methods. While these SOTA methods may align with one or more of the continual learning frameworks discussed earlier, they also have distinct continual learning strategies. We highlighted these differences from the generic continual learning frameworks in \textbf{Sec. \ref{subsec:sotabaselines}}. 

In this experiment, we incorporate STCR into three SOTA continual learning methods (\textbf{Sec. \ref{subsec:sotabaselines}}) and assess their generalization and continual learning performances across three image datasets, given two task sequence lengths of $T=6$ and $T=11$. The versions of the SOTA methods integrated with our STCR are labeled as ``SOTA + STCR". To facilitate comparisons, we also include the outcomes of their standalone versions without STCR. We present our results in \textbf{Tab. \ref{tab:shape-texture}}.

SOTA+STCR consistently outperforms SOTA themselves in terms of generalization and reduced forgetting.
For instance, when STCR is applied to AFC on ImageNet-100 with  $T$=$6$, AFC + STCR increases Acc. from 76.46\% to 79.24\% and decreases $\mathcal{F}$ from 9.87\% to 7.43\%, dramatically improving the continual learning performance. Meanwhile, AFC + STCR reduces $\mathcal{R}$-C by 15.67\% and improves $\mathcal{R}$-R by 8.83\%, demonstrating its effectiveness in enhancing the generalization ability of these SOTA continual learning methods. 
\begin{table*}[t!]
\centering
\footnotesize
\resizebox{1.90 \columnwidth}{!}{
  \begin{tabular}{@{}lccccccccccccccccccccc@{}}
    \toprule
    & \multicolumn{4}{c}{\textbf{ImageNet-100}} & \multicolumn{4}{c}{\textbf{ImageNet-1000}} & \multicolumn{3}{c}{\textbf{CIFAR-100}}\\
    \cmidrule(lr){2-5} \cmidrule(lr){6-9} \cmidrule(lr){10-12}
     &$\mathcal{R}$-C$\downarrow$ & $\mathcal{R}$-R $\uparrow$ & Acc. $\uparrow$ & $\mathcal{F}$ $\downarrow$  & $\mathcal{R}$-C$\downarrow$ & $\mathcal{R}$-R $\uparrow$ & Acc. $\uparrow$ & $\mathcal{F}$ $\downarrow$ & $\mathcal{R}$-C$\downarrow$ & Acc. $\uparrow$ & $\mathcal{F}$ $\downarrow$ \\
    \hline 
    CCIL\cite{mittal2021essentials} ($T$=$6$)
    &\makecell{89.85 \\ \scriptsize{(1.91)}}&\makecell{24.16 \\ \scriptsize{(1.73)}}&\makecell{75.30 \\ \scriptsize{(2.12)}}&\makecell{15.68 \\ \scriptsize{(2.32)}}
    &\makecell{95.73 \\ \scriptsize{(1.52)}}&\makecell{16.29 \\ \scriptsize{(2.18)}}&\makecell{66.92 \\ \scriptsize{(2.75)}}&\makecell{13.36 \\ \scriptsize{(1.94)}}
    &\makecell{67.81 \\ \scriptsize{(2.06)}}&\makecell{62.41 \\ \scriptsize{(1.91)}}&\makecell{19.35 \\ \scriptsize{(1.08)}}\\
    \rowcolor{Gray}
    CCIL + STCR ($T$=$6$, \textit{ours})
    &\makecell{\textbf{69.88} \\ \scriptsize{(2.71)}}&\makecell{\textbf{40.41} \\ \scriptsize{(3.04)}}&\makecell{\textbf{78.96} \\ \scriptsize{(1.73)}}&\makecell{\textbf{6.77} \\ \scriptsize{(0.48)}}
    &\makecell{\textbf{81.84} \\ \scriptsize{(2.31)}}&\makecell{\textbf{21.23} \\ \scriptsize{(1.67)}}&\makecell{\textbf{67.39} \\ \scriptsize{(0.89)}}&\makecell{\textbf{11.61} \\ \scriptsize{(2.45)}}
    &\makecell{\textbf{62.14} \\ \scriptsize{(2.96)}}&\makecell{\textbf{66.58} \\ \scriptsize{(2.92)}}&\makecell{\textbf{12.68} \\ \scriptsize{(3.80)}}\\
    Podnet \cite{douillard2020podnet} ($T$=$6$)
    &\makecell{111.18 \\ \scriptsize{(8.17)}}&\makecell{20.83 \\ \scriptsize{(2.99)}}&\makecell{75.65 \\ \scriptsize{(1.67)}}&\makecell{12.43 \\ \scriptsize{(1.10)}}
    &\makecell{116.52 \\ \scriptsize{(1.83)}}&\makecell{12.76 \\ \scriptsize{(2.92)}}&\makecell{67.14 \\ \scriptsize{(1.36)}}&\makecell{9.45 \\ \scriptsize{(2.58)}}
    &\makecell{68.87 \\ \scriptsize{(4.28)}}&\makecell{64.84 \\ \scriptsize{(0.29)}}&\makecell{15.08 \\ \scriptsize{(2.42)}}\\
    \rowcolor{Gray}
    Podnet + STCR ($T$=$6$, \textit{ours}) 
    &\makecell{\textbf{102.65} \\ \scriptsize{(1.26)}}&\makecell{\textbf{27.19} \\ \scriptsize{(2.94)}}&\makecell{\textbf{77.18} \\ \scriptsize{(1.18)}}&\makecell{\textbf{8.33} \\ \scriptsize{(1.11)}}
    &\makecell{\textbf{89.95} \\ \scriptsize{(1.24)}}&\makecell{\textbf{15.47} \\ \scriptsize{(0.76)}}&\makecell{\textbf{68.82} \\ \scriptsize{(2.13)}}&\makecell{\textbf{8.93} \\ \scriptsize{(1.62)}}
    &\makecell{\textbf{65.54} \\ \scriptsize{(3.91)}}&\makecell{\textbf{65.51} \\ \scriptsize{(1.23)}}&\makecell{\textbf{12.17} \\ \scriptsize{(2.30)}}\\
    AFC \cite{kang2022class} ($T$=$6$) 
    &\makecell{109.87 \\ \scriptsize{(0.95)}}&\makecell{18.83 \\ \scriptsize{(2.83)}}&\makecell{76.46 \\ \scriptsize{(1.03)}}&\makecell{9.87 \\ \scriptsize{(1.87)}}
    &\makecell{96.29 \\ \scriptsize{(2.76)}}&\makecell{12.82 \\ \scriptsize{(1.45)}}&\makecell{67.31 \\ \scriptsize{(0.94)}}&\makecell{8.27 \\ \scriptsize{(1.98)}}
    &\makecell{69.42 \\ \scriptsize{(2.51)}}&\makecell{63.85 \\ \scriptsize{(1.65)}}&\makecell{13.59 \\ \scriptsize{(1.09)}}\\
    \rowcolor{Gray}
    AFC + STCR ($T$=$6$, \textit{ours}) 
    &\makecell{\textbf{94.20} \\ \scriptsize{(3.39)}}&\makecell{\textbf{27.66} \\ \scriptsize{(1.02)}}&\makecell{\textbf{79.24} \\ \scriptsize{(1.18)}}&\makecell{\textbf{7.43} \\ \scriptsize{(1.50)}}
    &\makecell{\textbf{85.68} \\ \scriptsize{(2.34)}}&\makecell{\textbf{16.25} \\ \scriptsize{(1.87)}}&\makecell{\textbf{68.93} \\ \scriptsize{(1.52)}}&\makecell{\textbf{6.38} \\ \scriptsize{(0.85)}}
    &\makecell{\textbf{66.48} \\ \scriptsize{(3.38)}}&\makecell{\textbf{66.83} \\ \scriptsize{(1.85)}}&\makecell{\textbf{8.41} \\ \scriptsize{(3.41)}}\\
    \midrule
    CCIL\cite{mittal2021essentials} ($T$=$11$)
    &\makecell{93.80 \\ \scriptsize{(1.28)}}&\makecell{23.54 \\ \scriptsize{(2.55)}}&\makecell{72.16 \\ \scriptsize{(2.16)}}&\makecell{17.63 \\ \scriptsize{(1.19)}}
    &\makecell{96.84 \\ \scriptsize{(1.73)}}&\makecell{14.76 \\ \scriptsize{(2.28)}}&\makecell{64.73 \\ \scriptsize{(1.65)}}&\makecell{19.92 \\ \scriptsize{(2.82)}}
    &\makecell{71.51 \\ \scriptsize{(3.89)}}&\makecell{60.57 \\ \scriptsize{(3.10)}}&\makecell{22.77 \\ \scriptsize{(2.03)}}\\
    \rowcolor{Gray}
    CCIL + STCR ($T$=$11$, \textit{ours}) 
    &\makecell{\textbf{76.13} \\ \scriptsize{(4.04)}}&\makecell{\textbf{35.88} \\ \scriptsize{(2.19)}}&\makecell{\textbf{74.27} \\ \scriptsize{(0.86)}}&\makecell{\textbf{12.85} \\ \scriptsize{(0.63)}}
    &\makecell{\textbf{86.35} \\ \scriptsize{(0.96)}}&\makecell{\textbf{18.29} \\ \scriptsize{(1.34)}}&\makecell{\textbf{65.92} \\ \scriptsize{(2.47)}}&\makecell{\textbf{15.87} \\ \scriptsize{(1.59)}}
    &\makecell{\textbf{64.56} \\ \scriptsize{(3.84)}}&\makecell{\textbf{63.60} \\ \scriptsize{(1.54)}}&\makecell{\textbf{16.11} \\ \scriptsize{(3.16)}}\\
    Podnet\cite{douillard2020podnet} ($T$=$11$)
    &\makecell{117.85 \\ \scriptsize{(4.68)}}&\makecell{19.83 \\ \scriptsize{(2.54)}}&\makecell{73.16 \\ \scriptsize{(1.35)}}&\makecell{18.30 \\ \scriptsize{(0.90)}}
    &\makecell{118.27 \\ \scriptsize{(2.65)}}&\makecell{10.54 \\ \scriptsize{(1.78)}}&\makecell{64.41 \\ \scriptsize{(0.87)}}&\makecell{17.23 \\ \scriptsize{(2.16)}}
    &\makecell{71.07 \\ \scriptsize{(1.81)}}&\makecell{62.19 \\ \scriptsize{(1.68)}}&\makecell{21.44 \\ \scriptsize{(1.70)}}\\
    \rowcolor{Gray}
    Podnet + STCR ($T$=$11$, \textit{ours}) 
    &\makecell{\textbf{104.52} \\ \scriptsize{(4.75)}}&\makecell{\textbf{26.16} \\ \scriptsize{(2.57)}}&\makecell{\textbf{74.16} \\ \scriptsize{(1.43)}}&\makecell{\textbf{16.63} \\ \scriptsize{(1.76)}}
    &\makecell{\textbf{93.96} \\ \scriptsize{(1.42)}}&\makecell{\textbf{15.38} \\ \scriptsize{(2.53)}}&\makecell{\textbf{65.85} \\ \scriptsize{(1.93)}}&\makecell{\textbf{16.94} \\ \scriptsize{(0.78)}}
    &\makecell{\textbf{66.73} \\ \scriptsize{(2.76)}}&\makecell{\textbf{63.19} \\ \scriptsize{(0.73)}}&\makecell{\textbf{20.11} \\ \scriptsize{(1.65)}}\\
    AFC \cite{kang2022class} ($T$=$11$)
    &\makecell{113.20 \\ \scriptsize{(4.11)}}&\makecell{18.83 \\ \scriptsize{(1.54)}}&\makecell{74.49 \\ \scriptsize{(1.78)}}&\makecell{15.96 \\ \scriptsize{(1.22)}}
    &\makecell{89.61 \\ \scriptsize{(2.24)}}&\makecell{9.36 \\ \scriptsize{(1.15)}}&\makecell{66.58 \\ \scriptsize{(2.83)}}&\makecell{15.92 \\ \scriptsize{(1.37)}}
    &\makecell{70.18 \\ \scriptsize{(3.06)}}&\makecell{62.52 \\ \scriptsize{(0.98)}}&\makecell{16.43 \\ \scriptsize{(1.96)}}\\
    \rowcolor{Gray}
    AFC + STCR ($T$=$11$, \textit{ours})  
    &\makecell{\textbf{96.53} \\ \scriptsize{(3.80)}}&\makecell{\textbf{26.86} \\ \scriptsize{(3.32)}}&\makecell{\textbf{76.42} \\ \scriptsize{(1.97)}}&\makecell{\textbf{12.50} \\ \scriptsize{(2.21)}}
    &\makecell{\textbf{82.14} \\ \scriptsize{(2.97)}}&\makecell{\textbf{14.59} \\ \scriptsize{(1.82)}}&\makecell{\textbf{67.72} \\ \scriptsize{(0.95)}}&\makecell{\textbf{13.18}\\ \scriptsize{(2.39)}}
    &\makecell{\textbf{67.18} \\ \scriptsize{(3.43)}}&\makecell{\textbf{64.19} \\ \scriptsize{(0.95)}}&\makecell{\textbf{14.43} \\ \scriptsize{(0.53)}}\\
    \bottomrule
    \end{tabular}
}
\caption{\textbf{Generalization and continual learning performance of the state-of-the-art (SOTA) continual learning methods with and without our STCR.} We reported the performance of three SOTA continual learning methods with (highlighted in grey background) and without our STCR on two learning paradigms ($T$=$6$, the first six rows) and ($T$=$11$, the last six rows) across all three datasets (three columns) in terms of their generalization ability ($\mathcal{R}$-C and $R-R$) and continual learning ability (Acc., and $\mathcal{F}$). See \textbf{Sec. \ref{subsec:sotabaselines}} for the introduction of these SOTA methods. See \textbf{Sec. \ref{metrics}} for the introduction of the evaluation metrics. The best results are in bold. The numbers in brackets indicate the standard deviation after 3 runs. 
}
\label{tab:shape-texture}
\vspace{-10pt}
\end{table*}

\begin{figure}[t!]
\vspace{-4pt}
  \centering
  \includegraphics[width=0.9\columnwidth]{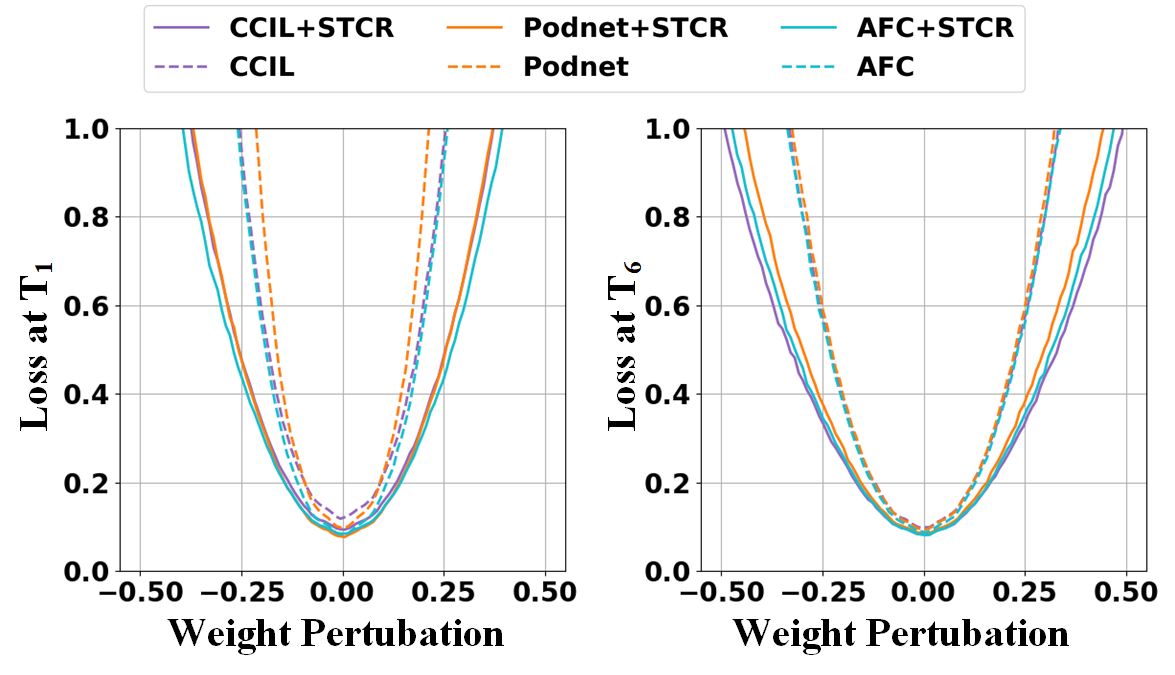}
  \vspace{-4pt}
  \caption{\textbf{Loss landscape analysis on existing continual learning methods with and without our STCR in the ImageNet-100 dataset.} 
  We perform the loss landscape analysis using the final model $\theta_6$ at task $\mathcal{T}_6$ with perturbed weights in random directions on the test sets of tasks $\mathcal{T}_1$ (left panel) and $\mathcal{T}_6$ (right panel) of the ImageNet-100 dataset. These final models are trained with competitive continual learning baselines with and without our STCR. See \textbf{Sec. \ref{subsec:sotabaselines}} for the introduction to continual learning baselines. 
  }
  \label{fig:landscape_stcr}
  \vspace{-15pt}
\end{figure}

\subsection{Continual learning methods with STCR have wider loss landscapes across tasks after weight perturbations}


In the previous subsection, SOTA continual learning methods with STCR demonstrated remarkable generalization and continual learning abilities across all four evaluation metrics (\textbf{Sec. \ref{metrics}}). Here, we investigate why STCR, as a generalization method, leads to less forgetting for SOTA continual learning methods. We applied the same experiment protocol and conducted loss landscape analysis, introduced in \textbf{Sec. \ref{subsec:lossland}}. In \textbf{Fig. \ref{fig:landscape_stcr}}, we present the loss landscapes of the models trained using SOTA continual learning methods with and without STCR. Consistent with observations in \textbf{Sec. \ref{subsec:lossland}}, we found that SOTA methods with STCR have wider loss landscapes compared to their counterparts (dotted versus solid lines in the same color in both test sets of $\mathcal{T}_1$ and $\mathcal{T}_6$). This suggests that models with STCR are more robust to weight perturbations, avoiding spurious features and enhancing the learning of generic features shared across tasks, thereby reducing catastrophic forgetting.



\subsection{Continual learning methods with STCR outperform their counterparts combined with existing generalization methods}
\label{stcrvsall}

We compare our STCR and the existing generalization baselines (\textbf{Sec. \ref{a-baselines}}) in a continual learning setting on ImageNet-100 with $T=6$. To isolate the effect of continual learning baselines, we use CCIL (\textbf{Sec. \ref{subsec:sotabaselines}}) as the continual learning backbone, apply all the generalization baselines to CCIL in every task, and evaluate their performances in $\mathcal{R}$-C, Acc., and $\mathcal{F}$ (\textbf{Sec. \ref{metrics}}). As controls, we also introduce the lower-bound, which is CCIL alone without any generalization methods. 

\begin{figure*}[t!]
\centering
\includegraphics[width=0.96\textwidth]{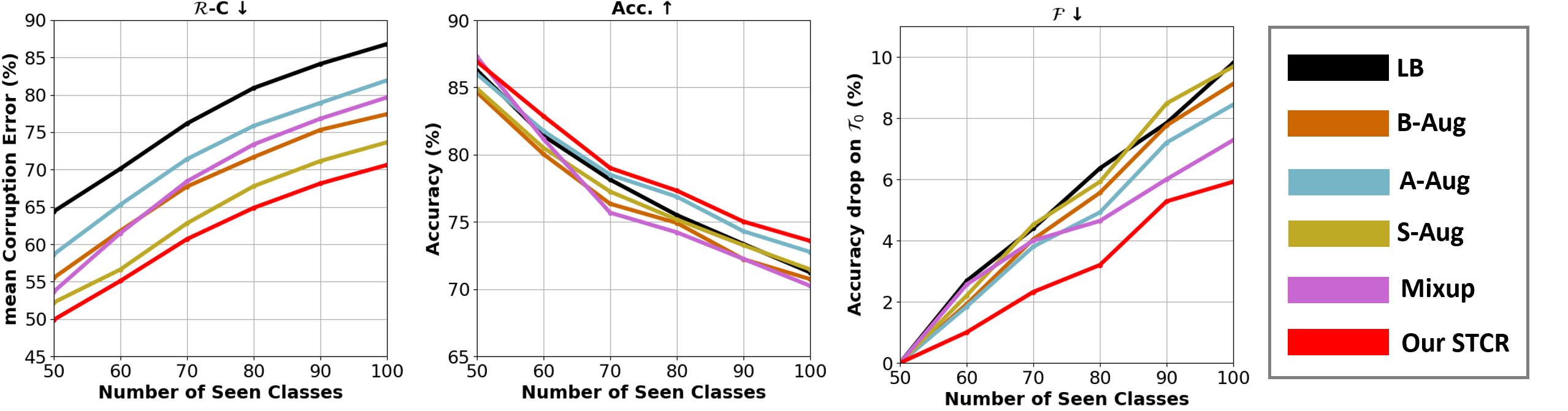}
\vspace{-4pt}
\caption{\textbf{The continual learning baseline with our STCR outperforms its counterpart with existing generalization methods.} We report the generalization ($\mathcal{R}$-C) and continual learning performances (Acc., and $\mathcal{F}$) of the continual learning method CCIL \cite{mittal2021essentials} integrated with our STCR as well as its counterpart integrated with established generalization baselines on ImageNet-100 with $T$=$6$. See \textbf{Sec. \ref{a-baselines}} for the introduction of each generalization baseline. See \textbf{Sec. \ref{metrics}} for the introduction of the evaluation metrics. The best performance is our STCR in red.
}
\label{fig:ablation}
\vspace{-10pt}
\end{figure*}

We report the results in \textbf{Fig. \ref{fig:ablation}}. As expected, the lower bound without any generalization methods performs the worst in terms of all the evaluation metrics. All the generalization baselines perform slightly better than the lower bound; however, their performance is still inferior to our STCR. This implies that our STCR is more effective in boosting the generalization ability within a task, and thereby enhancing the continual learning performance with reduced forgetting. 

It is also worth noting that S-Aug and Mixup share similarities with our STCR as both methods involve shape-texture conflict images. Even with the extra storage of artistic style templates, S-Aug still underperforms our STCR. This indicates that the data augmentation techniques alone are not enough. Additional regularization losses are necessary for the continual learning setting. 

Moreover, in contrast to Mixup loss which balances the learnt shape and texture biases, our STCR still achieves much better performance. This suggests that our proposed consistency regularization loss is more effective than Mixup for generalization in the continual learning setting.

\subsection{Network analysis reveals our key design decisions}
\label{sec:discussion}

%
\begin{table}[!ht]
\small
\centering
\resizebox{0.95\linewidth}{!}{
\begin{tabular}{llcccc}
    \toprule
    & & $\mathcal{R}$-C $\downarrow$ & $\mathcal{R}$-R $\uparrow$ & Acc. $\uparrow$ & $\mathcal{F}$ $\downarrow$ \\
    \midrule
    \multirow{4}[10]{*}{\begin{tabular}[c]{@{}l@{}}COEFFICIENT \\ $\gamma$\end{tabular}} 
    & $\gamma=0$ & \makecell{89.85 \\ \scriptsize{(1.91)}} & \makecell{24.16 \\ \scriptsize{(1.73)}} & \makecell{75.30 \\ \scriptsize{(2.12)}} & \makecell{15.68 \\ \scriptsize{(2.32)}} \\
    & $\gamma=0.001$ & \makecell{71.18 \\ \scriptsize{(3.27)}} & \makecell{37.22 \\ \scriptsize{(1.92)}} & \makecell{76.35 \\ \scriptsize{(1.23)}} & \makecell{9.20 \\ \scriptsize{(2.87)}} \\
    & $\gamma=0.01$ (ours) \cellcolor{Gray} & \makecell{\textbf{69.88} \\ \scriptsize{(2.71)}}\cellcolor{Gray} & \makecell{\textbf{40.41} \\ \scriptsize{(3.04)}}\cellcolor{Gray} & \makecell{\textbf{78.96} \\ \scriptsize{(1.73)}}\cellcolor{Gray} & \makecell{6.77 \\ \scriptsize{(0.48)}}\cellcolor{Gray} \\
    & $\gamma=0.1$ & \makecell{71.35 \\ \scriptsize{(2.28)}} & \makecell{39.19 \\ \scriptsize{(1.60)}} & \makecell{77.67 \\ \scriptsize{(1.42)}} & \makecell{\textbf{6.42} \\ \scriptsize{(1.80)}} \\
    \midrule

    \multirow{3}[7]{*}{\begin{tabular}[c]{@{}l@{}}COEFFICIENT \\ $\beta$\end{tabular}} 
    & $\beta=0.1$ & \makecell{102.80 \\ \scriptsize{(3.67)}} & \makecell{21.64 \\ \scriptsize{(2.45)}} & \makecell{59.85 \\ \scriptsize{(0.99)}} & \makecell{11.28 \\ \scriptsize{(0.58)}} \\
    & $\beta=1$ (ours) \cellcolor{Gray} & \makecell{\textbf{69.88} \\ \scriptsize{(2.71)}}\cellcolor{Gray} & \makecell{\textbf{40.41} \\ \scriptsize{(3.04)}}\cellcolor{Gray} & \makecell{\textbf{78.96} \\ \scriptsize{(1.73)}}\cellcolor{Gray} & \makecell{\textbf{6.77} \\ \scriptsize{(0.48)}}\cellcolor{Gray} \\
    & $\beta=10$  & \makecell{89.67 \\ \scriptsize{(3.05)}}  & \makecell{26.65 \\ \scriptsize{(1.25)}} & \makecell{70.18 \\ \scriptsize{(0.95)}}  & \makecell{13.28 \\ \scriptsize{(1.79)}}  \\
    \midrule

    \multirow{3}[7]{*}{TEMPERATURE} 
    & 1.0 & \makecell{73.06 \\ \scriptsize{(0.48)}} & \makecell{36.88 \\ \scriptsize{(1.57)}} & \makecell{72.37 \\ \scriptsize{(1.17)}} & \makecell{14.53 \\ \scriptsize{(2.91)}} \\
    & 2.0 (ours) \cellcolor{Gray} & \makecell{\textbf{69.88} \\ \scriptsize{(2.71)}}\cellcolor{Gray} & \makecell{\textbf{40.41} \\ \scriptsize{(3.04)}}\cellcolor{Gray} & \makecell{\textbf{78.96} \\ \scriptsize{(1.73)}}\cellcolor{Gray} & \makecell{\textbf{6.77} \\ \scriptsize{(0.48)}}\cellcolor{Gray} \\
    & 3.0 & \makecell{72.33 \\ \scriptsize{(1.50)}} & \makecell{38.41 \\ \scriptsize{(1.62)}} & \makecell{75.73 \\ \scriptsize{(1.28)}} & \makecell{9.63 \\ \scriptsize{(2.80)}} \\
    \midrule
  
    \multirow{3}[7]{*}{EXEMPLAR SIZE} 
    & 10 & \makecell{73.35 \\ \scriptsize{(0.82)}} & \makecell{36.05 \\ \scriptsize{(0.41)}} & \makecell{65.08 \\ \scriptsize{(0.71)}} & \makecell{11.35 \\ \scriptsize{(1.22)}} \\
    & 20 (ours) \cellcolor{Gray} & \makecell{\textbf{69.88} \\ \scriptsize{(2.71)}}\cellcolor{Gray} & \makecell{40.41 \\ \scriptsize{(2.04)}}\cellcolor{Gray} & \makecell{\textbf{78.96} \\ \scriptsize{(1.73)}}\cellcolor{Gray} & \makecell{6.77 \\ \scriptsize{(0.48)}}\cellcolor{Gray} \\
    & 30 & \makecell{69.14 \\ \scriptsize{(2.77)}} & \makecell{\textbf{40.66} \\ \scriptsize{(0.59)}} & \makecell{78.77 \\ \scriptsize{(0.53)}} & \makecell{\textbf{6.65} \\ \scriptsize{(0.16)}} \\
    \midrule
    
    \multirow{2}[5]{*}{REPLAY} 
    & Yes & \makecell{69.68 \\ \scriptsize{(2.50)}} & \makecell{37.45 \\ \scriptsize{(1.95)}} & \makecell{77.48 \\ \scriptsize{(2.46)}} & \makecell{7.92 \\ \scriptsize{(1.26)}} \\
    & No (ours) \cellcolor{Gray} & \makecell{\textbf{69.88} \\ \scriptsize{(2.71)}}\cellcolor{Gray} & \makecell{\textbf{40.41} \\ \scriptsize{(3.04)}}\cellcolor{Gray} & \makecell{\textbf{78.96} \\ \scriptsize{(1.73)}}\cellcolor{Gray} & \makecell{\textbf{6.77} \\ \scriptsize{(0.48)}}\cellcolor{Gray} \\
    \midrule

    \multirow{2}[5]{*}{\begin{tabular}[c]{@{}l@{}}STYLES \\ FROM\end{tabular}} 
    & current training set & \makecell{72.17 \\ \scriptsize{(2.33)}} & \makecell{35.83 \\ \scriptsize{(1.03)}} & \makecell{76.26 \\ \scriptsize{(1.64)}} & \makecell{8.94 \\ \scriptsize{(2.52)}} \\
    & exemplar sets (ours) \cellcolor{Gray} & \makecell{\textbf{69.88} \\ \scriptsize{(2.71)}}\cellcolor{Gray} & \makecell{\textbf{40.41} \\ \scriptsize{(3.04)}}\cellcolor{Gray} & \makecell{\textbf{78.96} \\ \scriptsize{(1.73)}}\cellcolor{Gray} & \makecell{\textbf{6.77} \\ \scriptsize{(0.48)}}\cellcolor{Gray} \\
    \midrule

    \multirow{3}[7]{*}{COLOR TRANSFER} 
    & CCIL & \makecell{89.85 \\ \scriptsize{(1.91)}} & \makecell{24.16 \\ \scriptsize{(1.73)}}  & \makecell{75.30 \\ \scriptsize{(2.12)}} & \makecell{15.68 \\ \scriptsize{(2.32)}} \\
    & w/ Color Transfer & \makecell{75.51 \\ \scriptsize{(1.12)}}  & \makecell{33.43 \\ \scriptsize{(1.97)}}  & \makecell{76.22 \\ \scriptsize{(1.50)}}  & \makecell{12.92 \\ \scriptsize{(1.75)}}  \\
    & w/ AdaIN (ours) \cellcolor{Gray} & \makecell{\textbf{69.88} \\ \scriptsize{(2.71)}}\cellcolor{Gray} & \makecell{\textbf{40.41} \\ \scriptsize{(3.04)}}\cellcolor{Gray} & \makecell{\textbf{78.96} \\ \scriptsize{(1.73)}}\cellcolor{Gray} & \makecell{\textbf{6.77} \\ \scriptsize{(0.48)}}\cellcolor{Gray} \\
    
    \midrule
    
    \multirow{4}[10]{*}{\begin{tabular}[c]{@{}l@{}}TASK \\ LENGTH\end{tabular}} 
    & CCIL \cite{mittal2021essentials} (T=26) & \makecell{109.13 \\ \scriptsize{(3.89)}} & \makecell{19.75 \\ \scriptsize{(0.96)}} & \makecell{66.47 \\ \scriptsize{(1.52)}} & \makecell{23.40 \\ \scriptsize{(0.74)}} \\
    & w/ STCR (T=26, ours) \cellcolor{Gray} & \makecell{\textbf{86.30} \\ \scriptsize{(1.89)}} \cellcolor{Gray} & \makecell{\textbf{28.94} \\ \scriptsize{(1.26)}} \cellcolor{Gray} & \makecell{\textbf{69.95} \\ \scriptsize{(1.79)}} \cellcolor{Gray} & \makecell{\textbf{21.01} \\ \scriptsize{(2.21)}} \cellcolor{Gray} \\
    & CCIL \cite{mittal2021essentials} (T=51) & \makecell{115.10 \\ \scriptsize{(2.65)}} & \makecell{16.40 \\ \scriptsize{(0.81)}} & \makecell{59.04 \\ \scriptsize{(0.98)}} & \makecell{35.54 \\ \scriptsize{(1.05)}} \\
    & w/ STCR (T=51, ours) \cellcolor{Gray} & \makecell{\textbf{99.40} \\ \scriptsize{(2.67)}} \cellcolor{Gray} & \makecell{\textbf{22.28} \\ \scriptsize{(1.58)}} \cellcolor{Gray} & \makecell{\textbf{62.16} \\ \scriptsize{(0.62)}} \cellcolor{Gray} & \makecell{\textbf{31.76} \\ \scriptsize{(0.81)}} \cellcolor{Gray} \\
    \midrule
  
    \multirow{4}[10]{*}{MODEL} 
    & ViTIL \cite{yu2021improving} & \makecell{85.54 \\ \scriptsize{(2.00)}} & \makecell{32.56 \\ \scriptsize{(1.23)}} & \makecell{77.75 \\ \scriptsize{(2.02)}} & \makecell{7.64 \\ \scriptsize{(1.74)}} \\
    & w/ STCR (ours) \cellcolor{Gray} & \makecell{\textbf{75.50} \\ \scriptsize{(2.27)}} \cellcolor{Gray} & \makecell{\textbf{37.92} \\ \scriptsize{(2.03)}} \cellcolor{Gray} & \makecell{\textbf{77.95} \\ \scriptsize{(0.93)}} \cellcolor{Gray} & \makecell{\textbf{6.82} \\ \scriptsize{(1.59)}} \cellcolor{Gray} \\
    & CCIL \cite{mittal2021essentials} & \makecell{89.85 \\ \scriptsize{(1.91)}} & \makecell{24.16 \\ \scriptsize{(1.73)}} & \makecell{75.30 \\ \scriptsize{(2.12)}} & \makecell{15.68 \\ \scriptsize{(2.32)}} \\
    & w/ STCR (ours) \cellcolor{Gray} & \makecell{\textbf{69.88} \\ \scriptsize{(2.71)}}\cellcolor{Gray} & \makecell{\textbf{40.41} \\ \scriptsize{(3.04)}}\cellcolor{Gray} & \makecell{\textbf{78.96} \\ \scriptsize{(1.73)}}\cellcolor{Gray} & \makecell{\textbf{6.77} \\ \scriptsize{(0.48)}}\cellcolor{Gray} \\
    \midrule
  
    \multirow{2}[5]{*}{MIXUP} 
    & Yes & \makecell{104.98 \\ \scriptsize{(2.76)}} & \makecell{20.74 \\ \scriptsize{(3.79)}} & \makecell{57.15 \\ \scriptsize{(0.99)}} & \makecell{10.92 \\ \scriptsize{(0.85)}} \\
    & No (ours) \cellcolor{Gray} & \makecell{\textbf{69.88} \\ \scriptsize{(2.71)}}\cellcolor{Gray} & \makecell{\textbf{40.41} \\ \scriptsize{(3.04)}}\cellcolor{Gray} & \makecell{\textbf{78.96} \\ \scriptsize{(1.73)}}\cellcolor{Gray} & \makecell{\textbf{6.77} \\ \scriptsize{(0.48)}}\cellcolor{Gray} \\
    \midrule
  
    \multirow{3}[7]{*}{\begin{tabular}[c]{@{}l@{}}10 TASKS \\10 CLASSES\end{tabular}} 
    & CCIL & \makecell{97.44 \\ \scriptsize{(3.62)}} & \makecell{21.41 \\ \scriptsize{(1.68)}} & \makecell{66.87 \\ \scriptsize{(1.53)}} & \makecell{20.71 \\ \scriptsize{(1.36)}} \\
    & CCIL+STCR (ours) \cellcolor{Gray} & \makecell{\textbf{78.78} \\ \scriptsize{(7.11)}}\cellcolor{Gray} & \makecell{\textbf{28.91} \\ \scriptsize{(1.61)}}\cellcolor{Gray} & \makecell{\textbf{70.19} \\ \scriptsize{(1.33)}}\cellcolor{Gray} & \makecell{14.69 \\ \scriptsize{(2.34)}}\cellcolor{Gray} \\
    & AFC+STCR (ours)  & \makecell{90.76 \\ \scriptsize{(4.05)}}  & \makecell{26.75 \\ \scriptsize{(1.44)}} & \makecell{70.17 \\ \scriptsize{(0.95)}}  & \makecell{\textbf{13.28} \\ \scriptsize{(1.79)}}  \\
    
    \bottomrule
  \end{tabular}

}
\caption{\textbf{Generalization and continual learning performance of our STCR variations in diverse application scenarios.} We reported the generalization ability ($\mathcal{R}$-C and $R-R$) and continual learning ability (Acc., and $\mathcal{F}$) of our STCR variations in multiple scenarios on ImageNet-100. 
See \textbf{Sec. \ref{sec:discussion}} for the introduction of each application scenario. 
See \textbf{Sec. \ref{metrics}} for the introduction of the evaluation metrics. The rows highlighted in gray indicate our default method design. The best results are in bold. The numbers in brackets indicate the standard deviation after 3 runs. 
}
\label{tab:discussion}
\vspace{-10pt}
\end{table}

To evaluate the contribution of each component in our STCR method, we repeat the experiments on ImageNet-100 using the ablated versions of our method. 
As control experiments, we fix the continual learning baseline CCIL(\cite{mittal2021essentials}, \textbf{Sec.\ref{subsec:sotabaselines}}) and use it to integrate with our ablated versions. In these experiments, the number of replay exemplars in CCIL is set to 2000 by default, unless otherwise specified.
We also introduce additional application scenarios and demonstrate the benefits that our STCR can bring in these scenarios.

\textbf{The effect of coefficient $\gamma$.} 
We analyze the effect of $\gamma$, which controls the strength of $\mathcal{L}^{STCR}$ in 
\textbf{Eq. \ref{eq:loss}}, by varying its values $\gamma \in \{0,0.001,0.01,0.1\}$. Note that $\mathcal{L}^{STCR}$ is not in effect when $\gamma$ is set to $0$. The results (COEFFICIENT $\gamma$) in \textbf{Tab. \ref{tab:discussion}}, demonstrate that incorporating STCR with small $\gamma$
leads to improved performance compared to the case when $\gamma{=}0$, in all evaluation metrics. This suggests that the STCR is essential for enhancing the continual learning performance. We also observed that higher values of $\gamma$ lead to less forgetting but may deteriorate generalization ability. 

It is important to note that the impact of $\gamma$ remains consistent across different learning protocols and datasets. For all protocols and datasets, we have fixed the default value of $\gamma{=}0.01$.

\textbf{The effect of coefficient $\beta$.}
We analyze the effect of the coefficient $\beta$, which controls the strength of $\beta\mathcal{L}_{\mathcal{P}}^{CE}$ in \textbf{Eq. \ref{eq:loss}}, by varying its values $\beta \in \{0.1, 1, 10\}$. We reported their results in \textbf{Tab. \ref{tab:discussion}} (COEFFICIENT $\beta$). From the results, we observed $\beta$ of 1 yields the best performance. Either a small $\beta$ of 0.1 or a large $\beta$ of 10 leads to decreased performance. Intuitively, a higher beta emphasizes the effect of naive replays on original exemplars in the memory buffer, leading to weaker representation alignment on shape-texture conflict images, and hence, poorer generalization performances. Lower beta impairs the effect of exemplar rehearsals during replays leading to more catastrophic forgetting.

\textbf{Hyper-Parameter Temperature $\tau$.}
The hyper-parameter temperature in \textbf{Eq. \ref{eq:shape-texture}} controls the smoothness of a probability distribution. Typically, the higher the temperature, the smoother the distributions, leading to a lower KL divergence and a softer alignment between the distributions. 
To study the effect of temperature, we conducted an extra network analysis by varying the temperatures from 1, 2, to 3. We reported the results of these models with different temperatures in \textbf{Tab. \ref{tab:discussion}} (TEMPERATURE). A temperature of 2 leads to optimal performance, while the performance decreases with temperatures of 1 and 3. These results suggest that the selection of temperature is essential for learning generic features of shape-texture conflict images. Empirically, we set the temperature to be 2, consistent with \cite{hinton2015distilling}. 

\textbf{Exemplar Set Size.}
Following \cite{rebuffi2017icarl,mittal2021essentials}, we fixed the memory buffer sizes to be 20 exemplars per object class in all the previous experiments. Here, we added an extra ablation study where we varied the memory buffer size from 10, 20, to 30 on CCIL+STCR, and studied their impacts. From the results in \textbf{Tab. \ref{tab:discussion}} (EXAMPLAR SIZE), we observed that the generalization and continual learning performances increased from 10 to 20. This implies that larger buffer sizes increase the diversity of exemplars during replays; hence, enhancing the performance of CCIL+STCR. We also noticed that the performance saturates from 20 to 30. This implies that too large memory buffer sizes do not necessarily lead to optimal performance.



\textbf{STCR in replay.} 
We investigate the impact of STCR in replays and present our findings in (REPLAY) in \textbf{Tab. \ref{tab:discussion}}. As emphasized in \textbf{Sec. \ref{sec:stcr}}, performing style transfers on the replay images and training on such synthesized shape-texture conflict images during replays hinder the performance of continual learning. The synthesized images may frequently exhibit low quality, potentially due to the restricted size of exemplars in the replay buffers. Rehearsing with these images could disrupt the retention of knowledge from earlier tasks.

\textbf{Styles from exemplar sets \textit{v.s.} current training set.}
For replay-based continual learning methods, we emphasized the importance of using style templates from exemplar sets in the memory buffer to generate shape-texture conflict images. Here, we report the performance when the style templates come from the current training set. The results in (STYLES FROM) in \textbf{Tab. \ref{tab:discussion}} show that using styles from exemplar sets outperforms the one from the current training set by around 3\% over the four evaluation metrics. Two possible reasons are: first, exemplar sets encompass a broader range of classes from previous tasks, enhancing the diversity of the shape-texture conflict images. Second, these conflict images carry styles from earlier tasks, serving as a form of feature rehearsal to mitigate forgetting.

\textbf{Color Transfer.} In our work, we used the style transfer method AdaIN \cite{huang2017arbitrary} as a proof-of-concept for generating shape-style conflict images. However, running AdaIN can be slow and compute-intensive for real-world practice.  
To verify that our STCR can be adapted to other types of style or color transfer methods, we designed a model variation by replacing AdaIn with the color transfer method \cite{reinhard2001color}. Color transfer is a process in digital image manipulation where the color palette of one image is applied to another, aiming to preserve the original content while changing its aesthetic and mood. From \textbf{Tab. \ref{tab:discussion}} (COLOR TRANSFER), we observed that the results of color transfers perform slightly worse than the ones with style transfers (AdaIN). This implies that color transfer, as a data augmentation method, is not as effective as style augmentation at capturing generic image representations. Interestingly, we also noted that the results of color transfer are better than the ones of CCIL alone across all four evaluation metrics. This suggests that a generic data augmentation technique could improve the model generalization ability; hence, leading to less forgetting in continual learning.   

\textbf{Longer Task Sequences.}
We conducted additional experiments on ImageNet-100 to evaluate the effectiveness of our STCR with longer task sequences $T{=}26$ and $T{=}51$. We evaluated CCIL (\cite{mittal2021essentials}, \textbf{Sec.\ref{subsec:sotabaselines}}) with and without our STCR in these two experiments. We presented the results in (TASK LENGTH) of \textbf{Tab. \ref{tab:discussion}}.
The results demonstrate that STCR can consistently improve both the generalization and continual learning performance of CCIL even over a much longer task sequence. For example, CCIL with STCR beats CCIL alone by 15.7\%, 6.88\% in $\mathcal{R}$-C and $\mathcal{R}$-R, and 3.12\% in Acc., and $\mathcal{F}$ over $T=51$.

\textbf{Connecting to vision transformer.} 
We assessed the effectiveness of our STCR on continual learning methods involving vision transformer architectures \cite{dosovitskiy2020image}. Specifically, we applied our STCR on Yu \etal \cite{yu2021improving} (ViTIL) for training vision transformers in continual learning settings.
We followed their implementation details and reported the results of their method with and without our STCR in (MODEL) in \textbf{Tab. \ref{tab:discussion}}.
Consistent with the results obtained from the convolutional neural network (CNN)-based CL approaches such as CCIL \cite{mittal2021essentials} (see (MODEL) in \textbf{Tab. \ref{tab:discussion}}), we observed improved performances of ViTIL in both generalization and continual learning evaluation metrics. Interestingly, we also noted that the performance boost in vision transformers is relatively lower than CNNs. 
This may be attributed to the fact that vision transformers have already learned better shape and texture representations than CNNs \cite{tuli2021convolutional}.

\textbf{Combining STCR with Mixup.}
We combined Shape-Texture Consistency Regularization (STCR) with MixUp using the losses proposed in \textbf{Eq. \ref{eq:loss}}. 
To achieve this, We followed the standard practice of MixUp operation \cite{zhang2017mixup} by using $\epsilon=0.5$
to blend a pair of the original training images in the current task. We then applied style transfers to these mixed-up images and used the proposed consistency regularization loss in \textbf{Eq. \ref{eq:shape-texture}} on their corresponding logits. Note that the logits of the mixed-up images are computed by taking the weighted sum of their logits given their class labels. 
We reported the result of this ablation study (MIXUP) in \textbf{Tab. \ref{tab:discussion}}. MixUp+STCR significantly underperforms STCR alone by 35.1\%, 19.67\%, 21.81\%, and 4.15\% in $\mathcal{R}$-C, $\mathcal{R}$-R, Acc, and F respectively. This suggests that the mixed images may interfere with the model's ability to maintain consistent shape and texture representations, which are critical for mitigating forgetting and enhancing generalization.

\textbf{Evaluation in a protocol of 10 tasks with every 10 classes per task.}
We also tested our model's performance using a different continual learning protocol, which prescribes a uniform number of classes per task. Specifically, we evaluated the competitive continual learning baselines CCIL and AFC, as well as their counterparts with our STCR (CCIL+STCR and AFC+STCR) on the ImageNet-100 dataset, where each task has 10 classes and there are a total of 10 tasks.
The results can be seen in \textbf{Tab. \ref{tab:discussion}} (10 TASKS, 10 CLASSES). We observed that all models performed worse than those in the previous protocol, where the initial task contained 50 classes and each subsequent task had 10 classes. This could be attributed to the increasing difficulty of the protocol.
Moreover, we found that CCIL+STCR outperforms CCIL alone in this new protocol, suggesting that our STCR consistently enhances the performance of existing continual learning methods by improving generalization and reducing forgetting.
Interestingly, we also compared the relative performance differences between AFC+STCR and CCIL+STCR in the current protocol with those in the previous protocol and found that the relative performance differences remain consistent. This indicates that our conclusions are independent of the protocol variation, as our STCR consistently benefits continual learning baselines across all evaluation metrics.


\section{Conclusion and future works}
\label{conlusionfuture}
In AI, the existing literature on generalization and continual learning has evolved independently. In our effort to bridge these fields, we presented empirical evidence showcasing their mutually beneficial relationship: effective generalization within a task facilitates quicker learning and improved performance in subsequent tasks within the continual learning framework. On the flip side, continual learning methods are designed to combat catastrophic forgetting, ensuring the preservation of knowledge from earlier tasks, and ultimately contributing to enhanced generalization for ongoing tasks. Building upon this insight, we introduced Shape-Texture Consistency Regularization (STCR), a simple yet effective regularization technique that learns both shape and texture representations for each task in continual learning. This approach integrated with any continual learning methods not only enhances generalization but also mitigates forgetting. Our extensive experiments highlight that the existing continual learning methods, seamlessly integrated with STCR, not only surpass their own performance but also outperform their counterparts integrated with any other existing generalization methods.


Despite the promising performance of our method, we have identified several limitations that need to be addressed in future work. First, as a proof-of-concept, our STCR uses a compute-heavy style transfer technique. For real-world applications, alternative style transfer methods that can perform few-shot fine-tuning and real-time inference should be explored. 

Second, we have only examined the interplay of generalization and continual learning in fully supervised settings. The dynamics of this interplay in semi-supervised and unsupervised settings remain to be explored.




\section*{Acknowledgements}
This research is supported by the National Research Foundation, Singapore under its AI Singapore Programme (AISG Award No: AISG2-RP-2021-025), and its NRFF award NRF-NRFF15-2023-0001. We also acknowledge Mengmi Zhang's Start Up Grant from Agency for Science, Technology, and Research (A*STAR), and Start Up Grant from Nanyang Technological University. Additionally, we acknowledge the support given to Zenglin Shi from the National Natural Science Foundation of China (62472138).


\bibliographystyle{IEEEtran}
\bibliography{ref}

\clearpage

\begin{IEEEbiography}[{\includegraphics[width=1in,height=1.25in,clip,keepaspectratio]{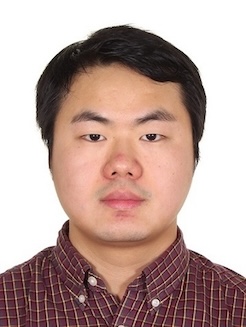}}]{Zenglin Shi} is a Full Professor at the Hefei University of Technology. He received his PhD degree from the University of Amsterdam in the Netherlands, where he worked closely with Prof. Cees Snoek and Dr. Pascal Mattes. After his PhD, He moved to Singapore to work as a research scientist at A*STAR. His research interests include Computer Vision and Machine Learning. He has published dozens of research papers in top-tier journals and conferences, such as TPAMI, IJCV, TIP, CVPR, ICCV.
\end{IEEEbiography}

\begin{IEEEbiography}[{\includegraphics[width=1in,height=1.25in,clip,keepaspectratio]{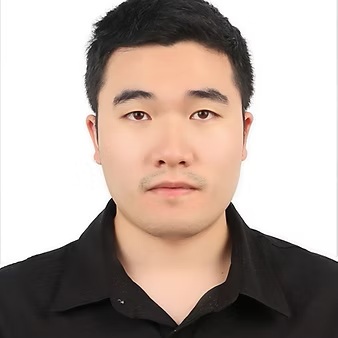}}]{Jie Jing} received the B.S. degree in Software Engineering from Sichuan University, Chengdu, China, in 2020. He is currently pursuing the Ph.D. degree with the Department of Computer Science, Sichuan University. He is also a visiting Ph.D. student with the Deep NeuroCognition Lab, Agency for Science, Technology and Research (A*STAR), Singapore. His research interests include computer vision, continual learning, and medical imaging.
\end{IEEEbiography}

\begin{IEEEbiography}[{\includegraphics[width=1in,height=1.25in,clip,keepaspectratio]{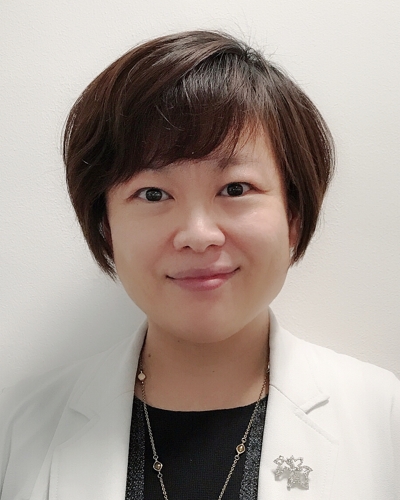}}]{Ying Sun} is currently a Principal Scientist at the Institute for Infocomm Research, Agency for Science, Technology and Research (A*STAR), Singapore. She received her B.Eng. from Tsinghua University, her M.Phil. from Hong Kong University of Science and Technology, and her Ph.D. in Electrical and Computer Engineering from Carnegie Mellon University. Her research interests include computer vision and machine learning, especially video understanding and generation, visual representation learning, and visual reasoning.
\end{IEEEbiography}

\begin{IEEEbiography}[{\includegraphics[width=1in,height=1.25in,clip,keepaspectratio]{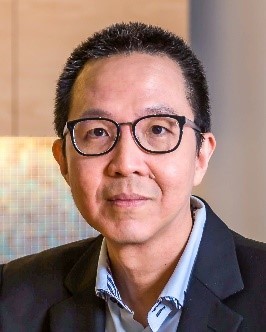}}]{Joo-Hwee Lim} received his B.Sc. (1st Class Honours) and M.Sc. (by research) degrees in Computer Science from the National University of Singapore and his Ph.D. degree in Computer Science \& Engineering from the University of New South Wales, Australia. He pioneered a novel framework (i.e. the popular bag-of-visual-words approach) for indexing images based on spatial aggregation of visual keywords, modeled from statistical learning, and a visual query language based on visual patterns, spatial quantifiers, and Boolean operators.  His research experience includes connectionist expert systems, neural-fuzzy systems, handwritten recognition, multi-agent systems, content-based image retrieval, scene/object recognition, medical image analysis, with >320 international refereed journal and conference papers, and 30 patents (awarded and pending). Dr. Lim is currently Senior Principal Scientist III and the Head of the Visual Intelligence Department at I2R, Singapore, and an Adjunct Professor at the College of Computing and Data Science (CCDS), Nanyang Technological University (NTU), Singapore. 
\end{IEEEbiography}

\begin{IEEEbiography}[{\includegraphics[width=1in,height=1.25in,clip,keepaspectratio]{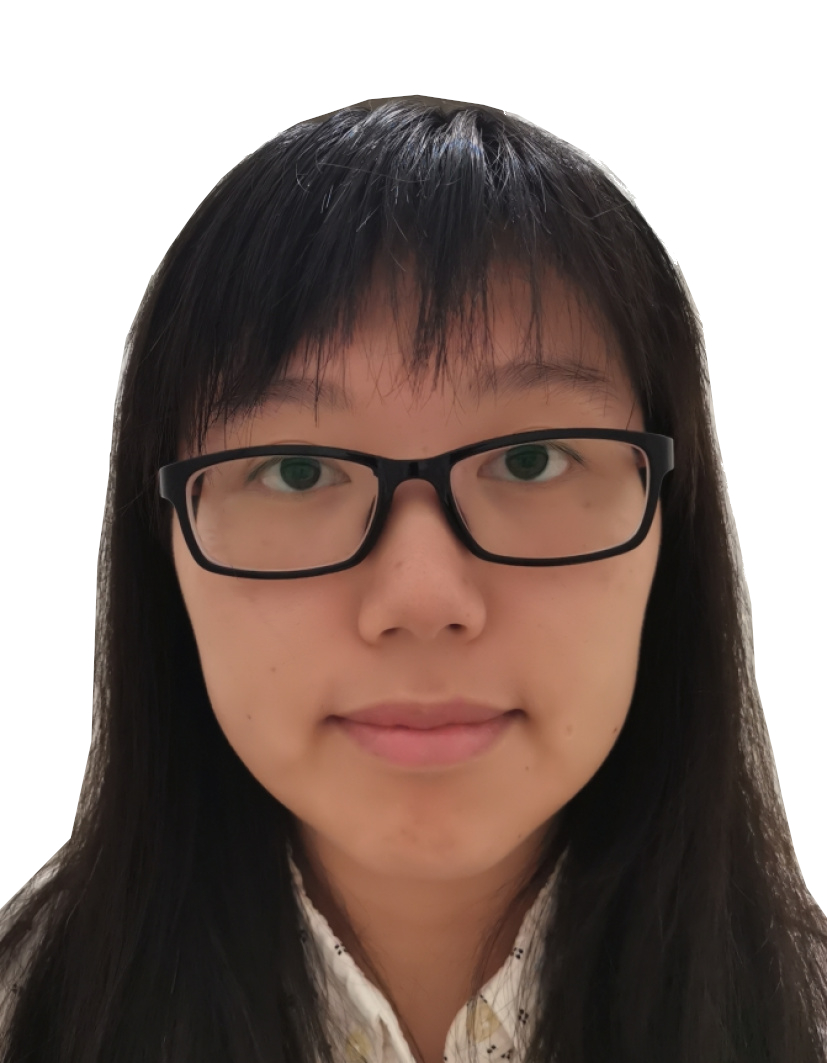}}]{Mengmi Zhang} is an Assistant Professor and Principal Investigator leading the Deep NeuroCognition Lab at Nanyang Technological University (NTU), Singapore. She also holds a joint appointment as a Principal Scientist at the Agency for Science, Technology and Research (A*STAR), Singapore. Her research lies at the intersection of artificial intelligence and computational neuroscience, with significant contributions to visual attention, contextual reasoning, semantic and episodic memory, and continual learning.
\end{IEEEbiography}

\vfill

\clearpage

\onecolumn
\renewcommand\appendixname{Supplementary Material}
\setcounter{page}{1} 
\begin{appendix}

\setcounter{table}{0}
\setcounter{figure}{0}
\setcounter{algocf}{0}
\counterwithout{table}{section}
\counterwithout{figure}{section}
\counterwithout{algocf}{section}

\renewcommand{\thesection}{S\arabic{section}}
\renewcommand{\thetable}{S\arabic{table}}
\renewcommand{\thefigure}{S\arabic{figure}}
\renewcommand{\thealgocf}{S\arabic{algocf}}

\subsection{Results with additional metrics}
\label{appendix:metrics}
We reported the results in average accuracy ($A_T$) and BWT for continual learning frames with and without STCR in \textbf{Tab. \ref{appendix:metricstable1}} and specific continual learning methods with and without STCR in \textbf{Tab. \ref{appendix:metricstable2}}. We found that continual learning frameworks and methods with our STCR outperform their counterparts by a significant margin in $A_T$ and BWT. This suggests that our STCR is effective at eliminating forgetting in continual learning. However, we also noted that the BWT results remain negative even with STCR. This suggests that the performance gap still exists in continual learning. 

\vspace{10pt}
\begin{multicols}{2}
\begin{table}[H]
    \centering
    \footnotesize
    \resizebox{\columnwidth}{!}{
    \begin{tabular}{@{}lcccccc@{}}
        \toprule
        \multirow{2}{*}{Method} & \multicolumn{2}{c}{$T$=6} & & \multicolumn{2}{c}{$T$=11} \\
        \cmidrule(lr){2-3} \cmidrule(lr){5-6}
         & $A_T(\%)\uparrow$ & $BWT\uparrow$ & & $A_T(\%)\uparrow$ & $BWT\uparrow$  \\
        \midrule
        Naive & \makecell{10.65 \\ \scriptsize{(0.14)}} & \makecell{-0.83 \\ \scriptsize{(0.05)}} & & \makecell{2.48 \\ \scriptsize{(1.41)}} & \makecell{-0.84 \\ \scriptsize{(0.09)}} \\
        \rowcolor{Gray} 
        Naive + STCR & \makecell{\textbf{11.85} \\ \scriptsize{(2.98)}} & \makecell{\textbf{-0.72} \\ \scriptsize{(0.07)}} & & \makecell{\textbf{9.48} \\ \scriptsize{(2.05)}} & \makecell{\textbf{-0.68} \\ \scriptsize{(0.16)}} \\
        \midrule
        WReg & \makecell{28.77 \\ \scriptsize{(2.39)}} & \makecell{-0.51 \\ \scriptsize{(0.12)}} & & \makecell{22.35 \\ \scriptsize{(4.11)}} & \makecell{-0.77 \\ \scriptsize{(0.07)}} \\
        \rowcolor{Gray} 
        WReg + STCR & \makecell{\textbf{41.07} \\ \scriptsize{(1.95)}} & \makecell{\textbf{-0.29} \\ \scriptsize{(0.03)}} & & \makecell{\textbf{35.07} \\ \scriptsize{(4.16)}} & \makecell{\textbf{-0.32} \\ \scriptsize{(0.05)}} \\
        \midrule
        Replay & \makecell{55.09 \\ \scriptsize{(1.09)}} & \makecell{-0.26 \\ \scriptsize{(0.01)}} & & \makecell{25.24 \\ \scriptsize{(8.02)}} & \makecell{-0.26 \\ \scriptsize{(0.02)}} \\
        \rowcolor{Gray} 
        Replay + STCR & \makecell{\textbf{58.24} \\ \scriptsize{(1.15)}} & \makecell{\textbf{-0.16} \\ \scriptsize{(0.01)}} & & \makecell{\textbf{45.89} \\ \scriptsize{(8.84)}} & \makecell{\textbf{-0.25} \\ \scriptsize{(0.04)}} \\
        \midrule
        KD & \makecell{55.65 \\ \scriptsize{(3.58)}} & \makecell{-0.16 \\ \scriptsize{(0.01)}} & & \makecell{35.81 \\ \scriptsize{(3.49)}} & \makecell{-0.15 \\ \scriptsize{(0.06)}} \\
        \rowcolor{Gray} 
        KD + STCR & \makecell{\textbf{61.18} \\ \scriptsize{(2.52)}} & \makecell{\textbf{-0.13} \\ \scriptsize{(0.02)}} & & \makecell{\textbf{48.05} \\ \scriptsize{(3.54)}} & \makecell{\textbf{-0.13} \\ \scriptsize{(0.02)}} \\
        \bottomrule
    \end{tabular}
    }
    \caption{
    \textbf{Continual learning performance of the generic continual learning frameworks with and without our STCR.} We reported the performance of four generic continual learning frameworks with (highlighted in gray background) and without our STCR on ImageNet-100 with $T$=$6$ (first column) and $T$=$11$ (second column) in terms of their continual learning ability (average accuracy ($A_T(\%)$) and backward transfer (BWT)). See \textbf{Sec. \ref{subsec:wkbaselines}} for the introduction of these generic continual learning frameworks. See \textbf{Sec. \ref{metrics}} for the introduction of the evaluation metrics. The best results are in bold. The numbers in brackets indicate the standard deviation after 3 runs. 
    }
\label{appendix:metricstable1}
\end{table}

\begin{table}[H]
    \centering
    \footnotesize
    \resizebox{\columnwidth}{!}{
    \begin{tabular}{@{}lcccccc@{}}
            \toprule
            & \multicolumn{3}{c}{\textbf{ImageNet-100}} & \multicolumn{2}{c}{\textbf{CIFAR-100}}\\
            \cmidrule(lr){2-4} \cmidrule(lr){5-6}
            & $A_T(\%)\uparrow$ & $BWT\uparrow$ & & $A_T(\%)\uparrow$ & $BWT\uparrow$   \\
            \hline 
            CCIL\cite{mittal2021essentials} ($T$=$6$)
            & \makecell{61.28 \\ \scriptsize{(0.38)}} & \makecell{-0.18 \\ \scriptsize{(0.03)}} & & \makecell{52.64 \\ \scriptsize{(0.49)}} & \makecell{-0.11 \\ \scriptsize{(0.03)}} \\
            \rowcolor{Gray}
            CCIL + STCR ($T$=$6$, \textit{ours})
            & \makecell{\textbf{75.82} \\ \scriptsize{(2.98)}} & \makecell{\textbf{-0.16} \\ \scriptsize{(0.07)}} & & \makecell{\textbf{58.03} \\ \scriptsize{(3.87)}} & \makecell{\textbf{-0.09} \\ \scriptsize{(0.06)}} \\
            Podnet \cite{douillard2020podnet} ($T$=$6$)
            & \makecell{63.97 \\ \scriptsize{(3.91)}} & \makecell{-0.16 \\ \scriptsize{(0.01)}} & & \makecell{52.19 \\ \scriptsize{(1.42)}} & \makecell{-0.06 \\ \scriptsize{(0.04)}} \\
            \rowcolor{Gray}
            Podnet + STCR ($T$=$6$, \textit{ours}) 
            & \makecell{\textbf{74.59} \\ \scriptsize{(1.23)}} & \makecell{\textbf{-0.15} \\ \scriptsize{(0.02)}} & & \makecell{\textbf{55.86} \\ \scriptsize{(1.34)}} & \makecell{\textbf{-0.05} \\ \scriptsize{(0.02)}} \\
            AFC \cite{kang2022class} ($T$=$6$) 
            & \makecell{65.74 \\ \scriptsize{(1.54)}} & \makecell{-0.09 \\ \scriptsize{(0.02)}} & & \makecell{53.76 \\ \scriptsize{(0.98)}} & \makecell{-0.06 \\ \scriptsize{(0.03)}} \\
            \rowcolor{Gray}
            AFC + STCR ($T$=$6$, \textit{ours}) 
            & \makecell{\textbf{75.39} \\ \scriptsize{(1.97)}} & \makecell{\textbf{-0.09} \\ \scriptsize{(0.05)}} & & \makecell{\textbf{55.08} \\ \scriptsize{(0.87)}} & \makecell{\textbf{-0.06} \\ \scriptsize{(0.03)}} \\
            \midrule
            CCIL\cite{mittal2021essentials} ($T$=$11$)
            & \makecell{58.79 \\ \scriptsize{(4.38)}} & \makecell{-0.15 \\ \scriptsize{(0.07)}} & & \makecell{47.69 \\ \scriptsize{(4.52)}} & \makecell{-0.09 \\ \scriptsize{(0.06)}} \\
            \rowcolor{Gray}
            CCIL + STCR ($T$=$11$, \textit{ours}) 
            & \makecell{\textbf{64.92} \\ \scriptsize{(1.95)}} & \makecell{\textbf{-0.13} \\ \scriptsize{(0.05)}} & & \makecell{\textbf{50.27} \\ \scriptsize{(0.87)}} & \makecell{\textbf{-0.03} \\ \scriptsize{(0.02)}} \\
            Podnet\cite{douillard2020podnet} ($T$=$11$)
            & \makecell{60.03 \\ \scriptsize{(4.43)}} & \makecell{-0.17 \\ \scriptsize{(0.03)}} & & \makecell{53.57 \\ \scriptsize{(0.83)}} & \makecell{-0.06 \\ \scriptsize{(0.04)}} \\
            \rowcolor{Gray}
            Podnet + STCR ($T$=$11$, \textit{ours}) 
            & \makecell{\textbf{61.78} \\ \scriptsize{(4.98)}} & \makecell{\textbf{-0.13} \\ \scriptsize{(0.04)}} & & \makecell{\textbf{53.29} \\ \scriptsize{(0.76)}} & \makecell{\textbf{-0.04} \\ \scriptsize{(0.02)}} \\
            AFC \cite{kang2022class} ($T$=$11$)
            & \makecell{61.84 \\ \scriptsize{(4.93)}} & \makecell{-0.16 \\ \scriptsize{(0.05)}} & & \makecell{56.94 \\ \scriptsize{(2.76)}} & \makecell{-0.07 \\ \scriptsize{(0.03)}} \\
            \rowcolor{Gray}
            AFC + STCR ($T$=$11$, \textit{ours})  
            & \makecell{\textbf{64.58} \\ \scriptsize{(4.38)}} & \makecell{\textbf{-0.12} \\ \scriptsize{(0.06)}} & & \makecell{\textbf{57.43} \\ \scriptsize{(2.47)}} & \makecell{\textbf{-0.03} \\ \scriptsize{(0.01)}} \\
            \bottomrule
            \end{tabular}
    }
    \caption{
    \textbf{Continual learning performance in average accuracy ($A_T(\%)$) and backward transfer (BWT) of the state-of-the-art (SOTA) continual learning methods with and without our STCR.} We reported the performance of three SOTA continual learning methods with (highlighted in grey background) and without our STCR on two learning paradigms ($T$=$6$, the first two columns) and ($T$=$11$, the last two columns) on ImageNet-100 and CIFAR-100 datasets in terms of their continual learning ability ($A_T(\%)$ and BWT). See \textbf{Sec. \ref{subsec:sotabaselines}} for the introduction of these SOTA methods. See \textbf{Sec. \ref{metrics}} for the introduction of the evaluation metrics. The best results are in bold. The numbers in brackets indicate the standard deviation after 3 runs.}
    \label{appendix:metricstable2}
\end{table}
\end{multicols}

\newpage
\subsection{Qualatative examples on style transferred images}
\label{appendix:styles}
\begin{figure*}[h]
  \centering
  \includegraphics[width=.99\textwidth]{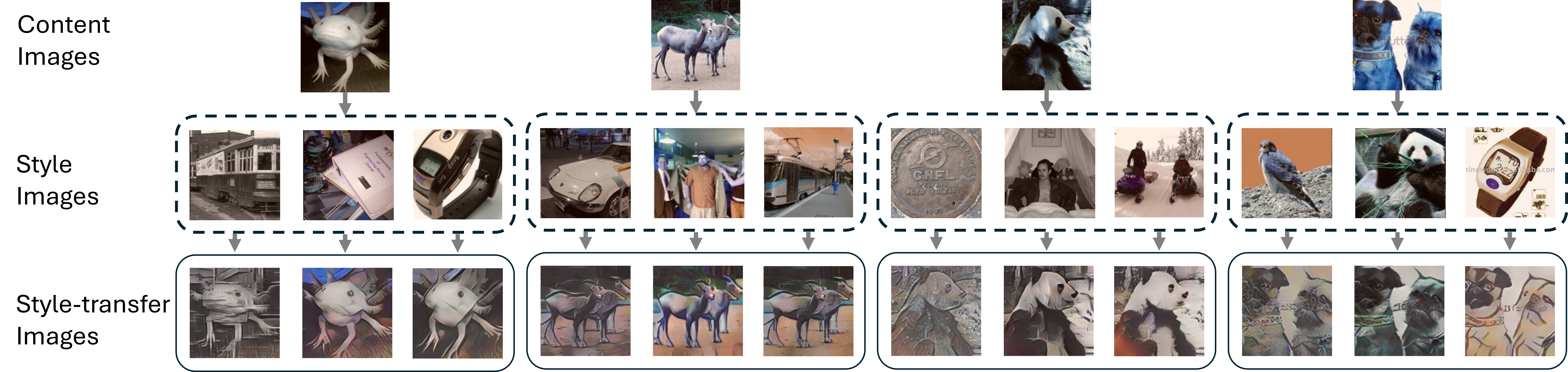}
  \caption{\textbf{Shape-texture conflict images created using AdaIn.} The top row displays the content images. The middle row features style images randomly selected from the exemplar set, and the bottom row shows the results of the style-transferred images. Three style images were randomly chosen for each content image, each with a corresponding style-transferred result. See \textbf{Sec. \ref{sec:std}} for details on generating shape-texture conflict images.
  }
  \label{appendix:stylesfig}
\end{figure*}

The quantitative evaluation of the generated shape-texture conflict images using AdaIn which has not been trained on the same dataset could be challenging. Here, we provide qualitative results visualizing the style-transferred images in \textbf{Fig. \ref{appendix:stylesfig}}. The source images are from the object classes of the current task, while the style images are randomly selected from the exemplars stored in the memory buffer. Note that the memory buffer only contains exemplar images from earlier tasks; hence, the source and style images are from different object classes. From these visualization results, we observed that the style-transferred images look reasonable. For example, in Column 3 and Row 3 of \textbf{Fig. \ref{appendix:stylesfig}}, the style transferred picture of a baby salamander has the style of watch but with the content of a baby salamander.

\subsection{Algorithum for STCR}
Here is the pseudo-code for our algorithm for STCR.
\begin{algorithm}[htb]
\caption{\footnotesize Integrate continual learning frameworks with STCR}
\label{algorithm}
  \SetKwData{Left}{left}\SetKwData{This}{this}\SetKwData{Up}{up}
  \SetKwFunction{Union}{Union}\SetKwFunction{FindCompress}{FindCompress}
  \SetKwInput{Input}{input}\SetKwInput{Require}{require}\SetKwInput{Output}{output}
 \Indm
  \Input{$\mathcal{D}_t$; /*{\footnotesize the training set for the new task }*/}
  \Require{$\mathcal{P}=\{\mathcal{P}_i\}_{i=1}^{t-1}$; /*{\footnotesize the current exemplar sets for replay}*/}
  \Require{$f$; /*{\footnotesize the style transfer model}*/}
  \Output{$\Theta_{t}$; /*{\footnotesize the new model to be learned from the new task}*/}
  \Indp
  $\Theta_{t} \leftarrow \Theta_{t-1}$; /*{\footnotesize add $m_t$ new output nodes}*/\\
  construct new exemplar set $\mathcal{P}_t$\;
  $\mathcal{P} \leftarrow \{\mathcal{P}_i\}_{i=1}^{t-1} \cup \mathcal{P}_t$ ; /*{\footnotesize add $\mathcal{P}_t$ to replay buffer}*/\\
  \For{iteration $j \leftarrow1$ \KwTo $l$}{
    load a mini-batch $(\mathcal{X}_t,\mathcal{Y}_t)$ from $\mathcal{D}_t$\;
    $\mathcal{X}_t \leftarrow \{(x_{i,t})\}_{i=1}^k$; $\mathcal{Y}_t \leftarrow \{(y_{i,t})\}_{i=1}^k$\;
    load a mini-batch $(\hat{\mathcal{X}}_t,\hat{\mathcal{Y}}_t)$ from $\mathcal{P}$\;
    $\hat{\mathcal{X}}_t \leftarrow \{(\hat{x}_{i,t})\}_{i=1}^k$; $\hat{\mathcal{Y}}_t \leftarrow \{(\hat{y}_{i,t})\}_{i=1}^k$\;
   $\tilde{\mathcal{X}}_t \leftarrow \{f(x_{i,t},\hat{x}_{i})\}_{i=1}^k$; /*{\footnotesize using Eq.(\ref{eq:style})}*/\\
    $\mathcal{Z}_t,\tilde{\mathcal{Z}}_t \leftarrow \Theta_t \big(\mathcal{X}_t,\tilde{\mathcal{X}}_t \big)$; /*{\footnotesize forward pass for new task}*/\\
    Compute $\mathcal{L}^{STCR}_{\mathcal{D}_t,\tilde{\mathcal{D}}_t}$ with $(\mathcal{Z}_t,\tilde{\mathcal{Z}}_t)$ using Eq.(\ref{eq:shape-texture})\;
    Compute $\mathcal{L}_{\mathcal{D}_t,\tilde{\mathcal{D}}_t}^{CE}$ with $\{(\mathcal{Z}_t,\mathcal{Y}_t),(\tilde{\mathcal{Z}}_t,\mathcal{Y}_t)\}$ \;
    $\hat{\mathcal{Z}}_t \leftarrow \Theta_{t} (\hat{\mathcal{X}})$; /*{\footnotesize forward pass for replay}*/\\
    Compute $\mathcal{L}_{\mathcal{P}}^{CE}$ with $(\hat{\mathcal{Z}}_t,\hat{\mathcal{Y}}_t)$ \;
    {\footnotesize $\mathcal{L}_t = \mathcal{L}_{\mathcal{D}_t,\tilde{\mathcal{D}}_t}^{CE}
   +\mathcal{L}_{\mathcal{P}}^{CE}+\gamma \mathcal{L}_{\mathcal{D}_t,\tilde{\mathcal{D}}_t}^{STCR}$}
   optimize $\Theta_{t}$ with loss $\mathcal{L}_t$; /*{\footnotesize backward pass}*/
  }

\end{algorithm}

\end{appendix}
\end{document}